    \PassOptionsToPackage{colorlinks=true}{hyperref}
    
\documentclass[preprint,12pt]{elsarticle}
    
   \usepackage[margin=1in]{geometry}
    
   \usepackage{amssymb}
\usepackage{amsmath}

\usepackage{amsthm}
\usepackage{booktabs}
\usepackage{enumitem}

\usepackage{float}
\usepackage{tikz}
\usetikzlibrary{shapes.geometric, arrows.meta, positioning}
\usepackage{graphicx}
\usepackage{multirow}
\usepackage{mathrsfs}
\usepackage[title]{appendix}
\usepackage{xcolor}
\usepackage{textcomp}
\usepackage{manyfoot}
\usepackage{algorithm}
\usepackage{subcaption}
\usepackage{algorithmicx}
\usepackage{algpseudocode}
\usepackage{listings}
\usepackage{comment}
\usepackage{hyperref}
\usepackage{cleveref}
\usepackage{orcidlink}
\usepackage{setspace}
\usepackage{lineno}
\usepackage[numbers]{natbib}
\bibpunct{[}{]}{,}{n}{,}{,}

    \journal{}
    
\begin{document}
    
    \begin{frontmatter}

   \title{MotoSafety: Edge-AI with Learned Temporal Importance for Two-Wheeler Collision Risk Assessment Under Time Pressure}

\author[inst1]{Sumit S. Shevtekar\orcidlink{0009-0003-7979-4607}\corref{cor1}\fnref{fn1}}
\author[inst1]{Chandresh K. Maurya\orcidlink{0000-0003-3519-600X}}
\author[inst2]{Gourab Sil\orcidlink{0000-0002-6930-4606}}
\author[inst3]{Subasish Das\orcidlink{0000-0002-1671-2753}}

\affiliation[inst1]{organization={Department of Computer Science and Engineering, Indian Institute of Technology Indore},
            addressline={Khandwa Road, Simrol},
            city={Indore},
            postcode={453552},
            state={Madhya Pradesh},
            country={India}}

\affiliation[inst2]{organization={Department of Civil Engineering, Indian Institute of Technology Indore},
            addressline={Khandwa Road, Simrol},
            city={Indore},
            postcode={453552},
            state={Madhya Pradesh},
            country={India}}

\affiliation[inst3]{organization={Civil Engineering Program, Ingram School of Engineering, Texas State University},
            addressline={RFM 5202},
            city={San Marcos},
            postcode={78666},
            state={TX},
            country={USA}}

\cortext[cor1]{Corresponding author. E-mail: sumit.shevtekar@gmail.com}
      
    
\begin{abstract}
Powered two-wheeler riders face critical safety challenges in low- and middle-income countries, yet limited studies exist on how cognitive stressors such as Time Pressure influence collision risk. We address this gap by introducing a comprehensive dataset consisting of over 129,000 labeled multivariate time-series feature windows, gathered across 153 simulator rides from 51 participants under No, Low, and High TP scenarios. Across each sequence, we capture 64 distinct attributes covering vehicle motion, rider control actions, spatial proximity, and rule compliance indicators. Using this dataset, we introduce MotoSafety, a new edge-AI framework built on the Learned Temporal Importance (LTI) concept. MotoSafety achieves 94.97\% accuracy and 99.33\% ROC AUC, outperforming ten baselines, including TimesNet and LLM4TS, and achieves 0.039 MSE and 0.094 MAE for forecasting (4.4$\times$ lower error than Time-LLM and iTransformer). With only 1.15M parameters and 0.135 ms latency, it is suitable for edge deployment on low-cost CPU hardware. Using ground truth TP as an inductive bias improves accuracy from 94.09\% to 94.97\%, while predicted TP achieves 94.82\%. Using only 21 IMU+GPS features, it achieves 93.91\% accuracy, indicating practical deployment. Beyond PTW safety, the architecture shows better transferability to human activity (97.66\%) and clinical (99.65\%) domains. This lightweight framework advances PTW collision risk assessment, supporting the Safe System Approach for Intelligent Transportation Systems.
\end{abstract}
    
    \begin{graphicalabstract}
    \centering
        \includegraphics[width=1\textwidth]{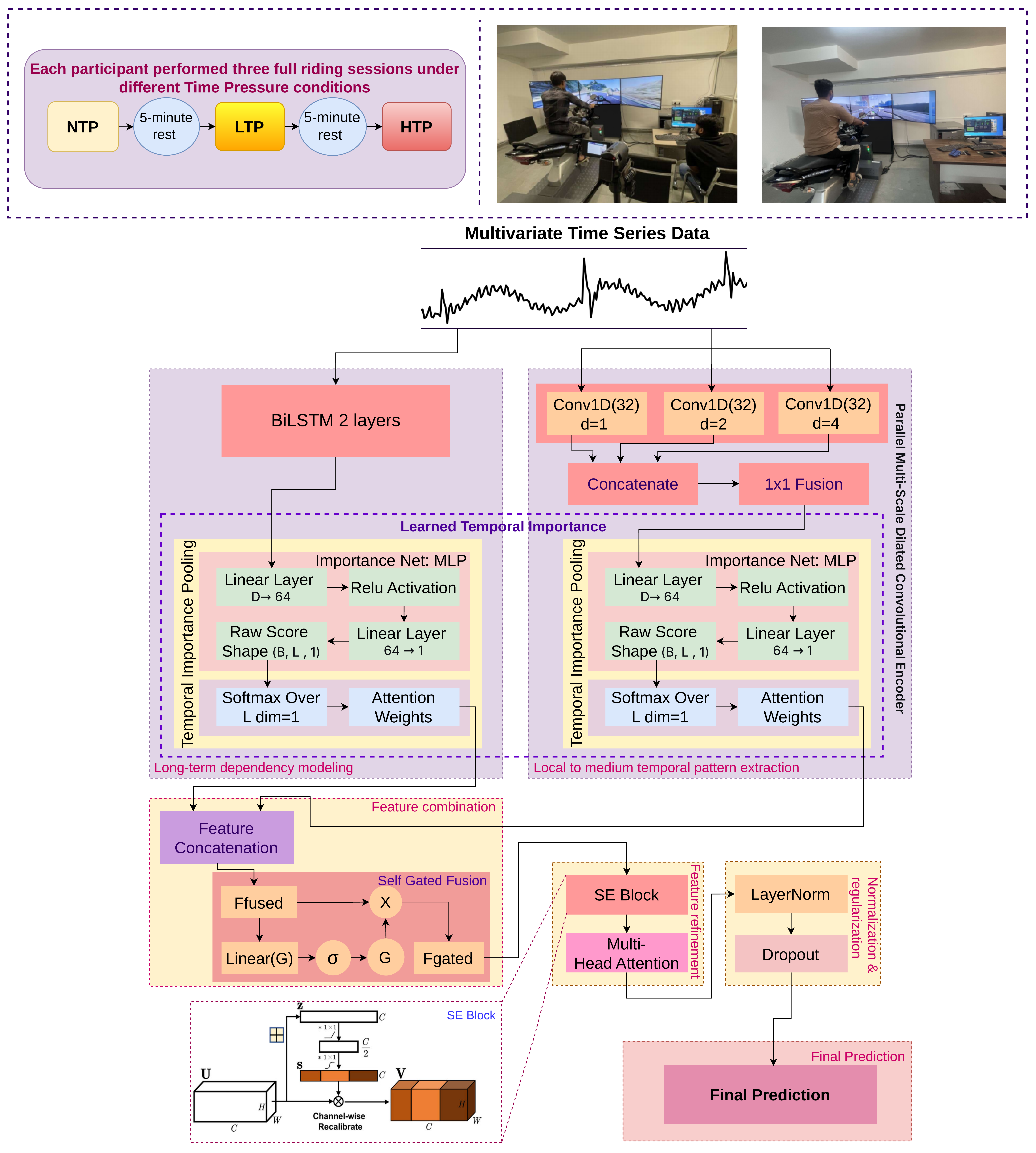}
    \end{graphicalabstract}

    
    

    \end{frontmatter}
    
    


    \section{Introduction}\label{sec1}
        
\begin{figure}  
    \centering
    \begin{minipage}{0.39\linewidth}
        \centering
        \includegraphics[width=\linewidth]{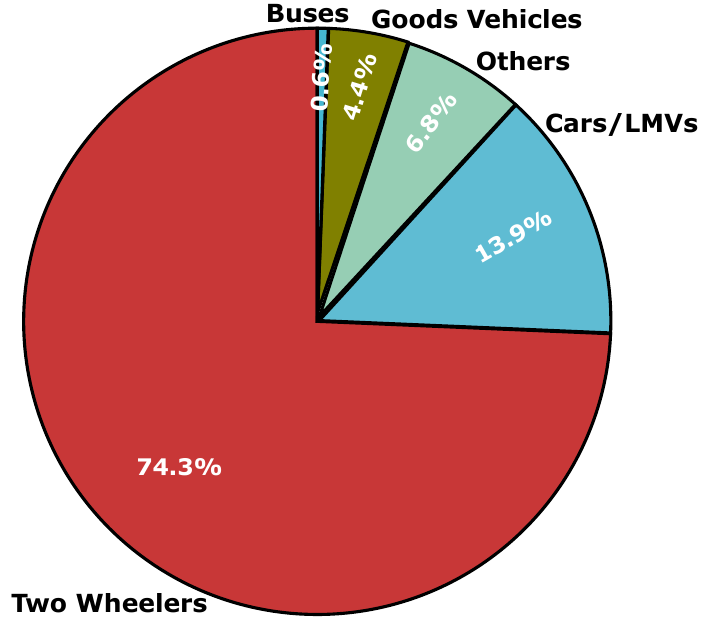}
        \caption{Registered Vehicles Share.}
        \label{fig:RegVehicle}
    \end{minipage}
        \hspace{0.04\linewidth}
    \begin{minipage}{0.45\linewidth}
        \centering
        \includegraphics[width=\linewidth]{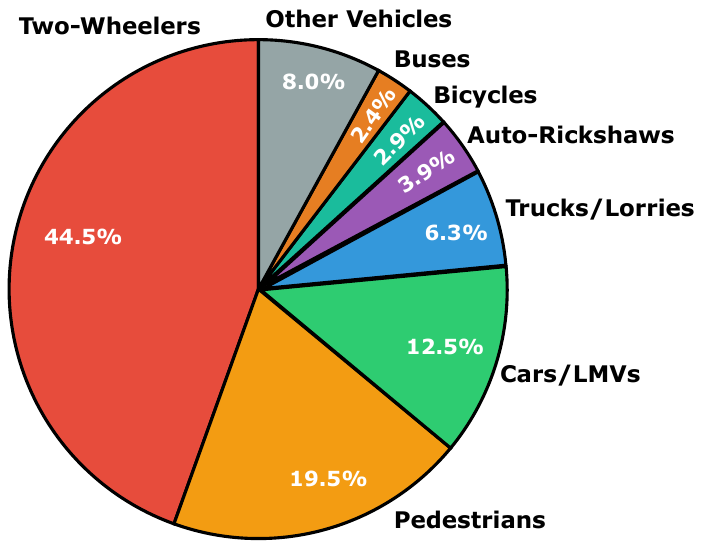}
        \caption{Road Crash Fatalities in India.}
        \label{fig:Vehicle_Fatalities}
    \end{minipage}
\end{figure}

Road traffic crashes are a major global health concern, particularly affecting Low- and Middle-Income Countries (LMICs), where the mortality burden is high~\citep{WHO2023}. Powered two-wheeler (PTW) riders represent one of the most vulnerable groups, facing disproportionately high fatality rates due to limited protection, complex traffic, and socio-economic pressures to meet mobility and livelihood deadlines. Globally, PTWs account for approximately 21\% of road traffic fatalities, a figure that rises to nearly 38\% in LMICs, where PTWs serve as an affordable and widely adopted mode of transport~\citep{WHO2023}. In India alone, PTWs constituted 74.4\% of registered vehicles as of 2022 shown in Fig.~\ref{fig:RegVehicle}~\citep{MORTH202324}. As of 2023, two-wheelers constitute the largest segment of the India’s vehicle population with more than 263 million registered units and were involved in 44.5\% of total road traffic fatalities, as shown in Fig.~\ref{fig:Vehicle_Fatalities}~\citep{MORTH202425,MORTH202223}.
Official MoRTH data from 2019 to 2023 consistently indicates that male riders account for the vast majority of road fatalities in India, with proportions varying between 85.2\% and 87.3\% across these years~\citep{morth2019,morth2020,morth2021,morth2022,morth2023}. The highest concentration of these fatalities is observed within the 18–45 age demographic. These patterns underscore the elevated vulnerability of male PTW riders in India's traffic environment. Despite advances in infrastructure design and traffic enforcement, human error remains the dominant contributor to PTW crashes, with risky behaviors such as overspeeding, abrupt maneuvering, and inadequate hazard response accounting for a large share of fatalities~\citep{Sharma21042025}. Naturalistic and simulator-based studies consistently show that riding speed, control variability, and situational context strongly influence crash risk~\citep{KONG2020105620}. The operating speeds of PTW riders are considerably higher than those of four-wheeler drivers, often reaching up to 2.3 times the speeds observed in cars~\citep{PAWAR2022105582}. 

Time Pressure (TP) refers to the subjective perception of insufficient time to complete a riding task, commonly arising from risk-prone behaviors, situational urgency, or externally imposed constraints~\citep{PAWAR2022105582,fortune2025quickcommerce}. This is especially common for app-based delivery riders in LMICs such as India. App-generated delivery deadlines, performance incentives, and intense customer demands for fast, flawless service combine to create persistent and often extreme TP~\citep{fortune2025quickcommerce,thehindu2022delivery,jakartapost2022delivery}. Elevated TP affects cognitive processing and motor control, narrowing attentional bandwidth and accelerating decision-making. This in turn promotes risk-prone behaviors such as overspeeding, inconsistent braking, and delayed hazard response~\citep{PAWAR2022105582,Gupta25112024,Sharma21042025}. Simulator studies on car drivers report increased accident risk of up to 181\% under high time pressure (HTP) compared to no time pressure (NTP)~\citep{PAWAR2022105582}. Related work \cite{pavlidis2016dissecting} demonstrates that cognitive and emotional stress amplifies speed variability and control instability. However, a key cause of these behaviors, TP-induced cognitive stress, is still not well understood within the PTW-driving situation. Recent AI-driven traffic crash prediction leverages large-scale datasets, classical ML/DL architectures to improve safety~\citep{Mostafa2025,Pan2024,Bouhsissin2024,Aci2025,Jiang2025}. Although effective in predicting four-wheeler crashes, these approaches inadequately model the unique dynamics and highly unstable control characteristics of PTWs. For example, \cite{Pan2024} achieved $\sim$90\% accuracy using deep neural networks on highway data, \cite{Bouhsissin2024} improved driver behavior classification with feature selection on large datasets, \cite{Aci2025} used Random Forest to predict injury severity (92\% accuracy). However, these studies focus primarily on \emph{four-wheelers or general traffic datasets and do not incorporate fine-grained temporal modeling of PTW riders or cognitive stress factors such as TP}.

To our knowledge, this is the first study to focus on predicting collision risk for two-wheeler riders under TP conditions. To address this gap, we introduce the \textbf{MotoSafety} specifically designed for multivariate time series data and to quantify and predict collision risk by modeling the interplay between a riders cognitive state under TP and their operational maneuvers. Our work addresses a critical public safety exigency: the protection of vulnerable PTW riders operating under chronic TP—a phenomenon increasingly prevalent in the rapidly urbanizing and dense traffic environments of emerging economies like India. The proposed system can be deployed as an edge-based alert system, where a handlebar-mounted device or smart helmet provides haptic/audio alerts when collision risk exceeds a threshold, enabling proactive safety interventions for riders under TP (e.g., emergency commuting, delivery deadlines, or urgent travel). 

This methodology was co-developed with road safety authorities and transportation experts to address PTW fatalities in LMICs. By integrating stakeholder input on crash-prone scenarios and demographic vulnerability, we ensure that our edge-deployable model satisfies real-world operational requirements. Consequently, our approach transitions from theoretical inference to a practitioner-validated solution for tangible public safety impact. The major contributions of this research are summarized as:

\subsection{Key Contributions}

We make the following key contributions.

\begin{enumerate}
\item \textbf{Large-Scale PTW Simulator Data Under TP:}
    To study the effect of TP on PTW riders, we collected  129,209 feature windows extracted via sliding window segmentation from 153 riding sessions (51 riders × 3 TP conditions).  Each input sequence comprises 64 features categorized into vehicle dynamics, control inputs, headway/tailway distances, and behavioral violations.

\item \textbf{Novel Architecture for Collision Risk Forecasting and Classification:} 
We propose MotoSafety, a novel architecture grounded in the Learned Temporal Importance (LTI) principle. The architecture integrates: (i) parallel multi-scale dilated convolutions for local-to-medium pattern extraction; (ii) Bidirectional Long Short-Term Memory (Bi-LSTM) for long-range dependencies; (iii) Temporal Importance Pooling (TIP) for content-aware temporal collapse, applied independently to both CNN and BiLSTM branches, which reduces downstream fusion, SE-recalibration, and attention stages to $\mathcal{O}(1)$ relative to the sequence length, while the front-end encoder operates at $\mathcal{O}(L)$ complexity, resulting in an overall $\mathcal{O}(L)$ framework (Table~\ref{tab:complexity}); and (iv) a self-gated fusion mechanism that leverages squeeze-and-excitation (SE) blocks and multi-head attention (MHA) for adaptive representation refinement.

\item \textbf{SOTA Performance and Deployability:} MotoSafety achieves 94.97\% accuracy and 99.33\% ROC AUC, showing improved performance over ten baselines including TimesNet, Time-LLM and LLM4TS. For long-term forecasting, it achieves 0.039 MSE and 0.094 MAE (4.4× lower error than Time-LLM and iTransformer). Notably, MotoSafety requires only $\sim$1.15M parameters and achieves an inference latency of 0.135~ms, making it 21.9$\times$ and 71.9$\times$ faster than TimesNet and LLM4TS, respectively, and makes it suitable for edge devices.

\item \textbf{TP Inductive Bias, Real-World Feasibility, and Architecture Transferability:} Explicit TP prediction improves collision accuracy from 94.09\% to 94.97\% (ground truth) and 94.82\% (predicted TP). Using only 21 real-world features (IMU+GPS), MotoSafety achieves 93.91\% accuracy, indicating practical deployment potential; hardware validation remains future work. Beyond PTW safety, the architecture demonstrates strong transferability to human activity (97.66\%) and clinical exercise (99.65\%) domains when retrained from scratch.

\end{enumerate}

\section{Related Work}
\label{sec:RelatedWork}

\subsection{Time Pressure as a Latent Risk Factor in Driving Safety}

Time Pressure (TP) is widely recognized as a cognitive stressor that degrades driver and rider decision-making and elevates crash risk. Under TP, individuals experience a perceived urgency to complete the driving task within insufficient time, which narrows attentional focus, accelerates cognitive processing, and increases reliance on risk-oriented heuristics~\citep{PAWAR2022105582,Gupta25112024}. Empirical studies consistently show that TP leads to higher speeds, shorter accepted gaps, reduced safety margins, and increased crash likelihood across diverse traffic contexts~\citep{PAWAR20221,Sharma21042025,fortune2025quickcommerce}. Experimental evidence from simulator-based studies indicates that TP significantly alters longitudinal and lateral control behavior, including increased acceleration variability, unstable steering, and abrupt braking~\citep{PAWAR202229}.

These behavioral deviations are strongly associated with near-crash and collision events, suggesting that TP acts as an upstream cognitive trigger for unsafe driving outcomes rather than merely a contextual modifier. PTWs are particularly susceptible to TP-induced risk due to unstable dynamics and high dependence on rider motor control compared to four-wheelers~\citep{PAWAR20221,Sharma21042025,GASHAW2018148}. At unsignalized intersections, drivers experiencing low time pressure (LTP) and high time pressure (HTP) tend to accept substantially shorter gaps, leading to crash probability increases of 127\% under LTP and 181\% under HTP compared to no time pressure (NTP) conditions~\citep{PAWAR2022105582}. TP increases cognitive load and unintentional attentional lapses under stress~\citep{GUPTA2022105820,Leung01112012}, reducing situational awareness and leading to errors that precede collisions. These findings highlight TP as a critical contributor to risk in dense, heterogeneous traffic flows common in LMICs.

\subsection{Simulator-Based Collision and Risk Analysis}

Driving simulators offer a secure and regulated setting for examining collision risk under cognitively demanding scenarios, including TP. They eliminate the hazards associated with real-world driving while facilitating high-resolution data acquisition for both collision risk classification and kinematic state prediction~\citep{PAWAR202229,Bham04032018,LI2019288}. Prior work demonstrates strong relative validity between simulated and real-world behavior, particularly for risk-related metrics including speed selection, braking intensity, and control variability~\citep{PAWAR202229,Bham04032018,LI2019288}. Although the magnitude of observed behaviors may differ between simulator and naturalistic conditions, the core behavioral responses to TP are preserved: both environments exhibit higher speeds, more frequent lane changes, stronger braking, and greater variability in vehicle control~\citep{PAWAR202229}. This supports the ecological validity of using simulators to investigate TP-induced behavioral changes in PTW rider behavior, enabling the current study to train and evaluate MotoSafety's classification and forecasting tasks under controlled yet realistic conditions.

\subsection{Machine Learning for Collision and Risk Prediction}

The application of ML and DL methods in intelligent transportation systems (ITS) has grown considerably, particularly for predicting risky events, collisions, and hazardous driving behaviors. Traditional techniques such as Decision Tree, Support Vector Machines, and feed-forward neural networks have shown moderate effectiveness in detecting high-risk maneuvers~\citep{Bouhsissin2023,Rodegast2024}. In recent years, sequence-based architectures like Temporal Transformers, Informer, and TimesNet have demonstrated considerable effectiveness in capturing long-range temporal patterns in driving behavior data~\citep{zhou2021informer,10.1609/aaai.v37i9.26317}. However, most existing models focus on detecting externally observable outcomes (e.g., collisions or near-crashes) after risk has already manifested. They largely overlook latent cognitive precursors such as TP that shape behavior well before unsafe actions occur. Furthermore, few studies address PTW-specific dynamics, which are characterized by inherent instability and high dependence on rider motor control.

\subsection{Research Gap and Motivation}

Although TP is a critical contributor to unsafe driving behavior, it is rarely integrated into collision risk assessment frameworks as a latent risk factor. This gap is particularly pronounced for PTWs, where safety margins are narrow and early intervention is essential. Motivated by these limitations, this work advances collision risk assessment under TP by leveraging high-resolution multivariate time-series data to infer risk directly from behavioral dynamics. By focusing on collision risk in cognitively stressed riding conditions, this study aligns with the Safe System Approach (SSA)~\citep{who_rti_2023} and the goal of Vision Zero~\citep{krug2022who}, enabling the detection of high-risk states and supporting the development of intelligent rider-assistance systems and enhanced safety for PTW users.

\section{Methodology}
In this section, we provide details of our data collection protocol, followed by MotoSafety architecture.
\subsection{Experimental Design and Data Collection}
\label{simulator}

\subsubsection{Simulator Setup}
\label{subsec:simulator_setup}

\begin{figure}[t]
    \centering
    \begin{subfigure}[b]{0.45\textwidth}
        \centering
        \includegraphics[width=\textwidth]{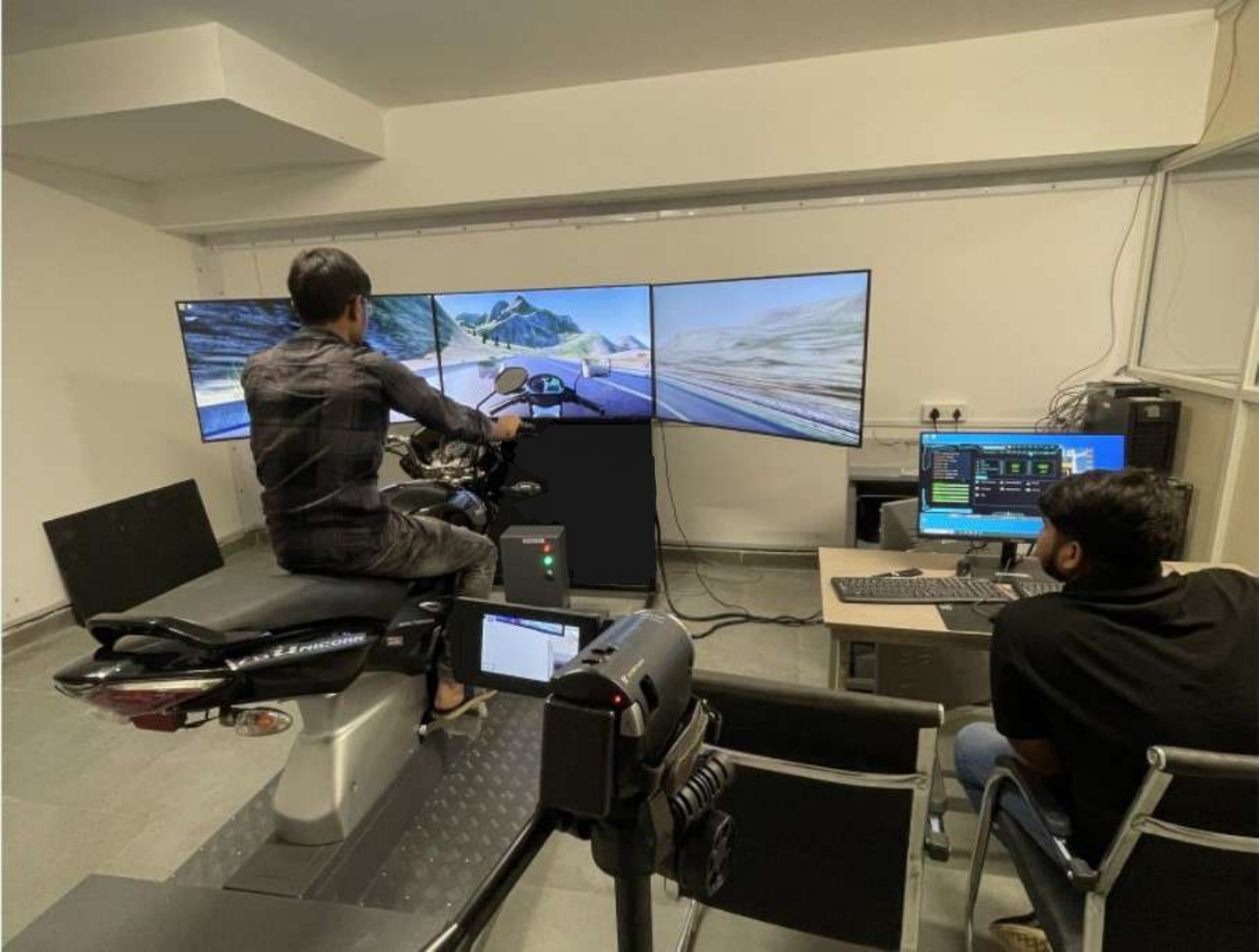}
        \label{fig:p1}
    \end{subfigure}
    \hfill
    \begin{subfigure}[b]{0.45\textwidth}
        \centering
        \includegraphics[width=\textwidth]{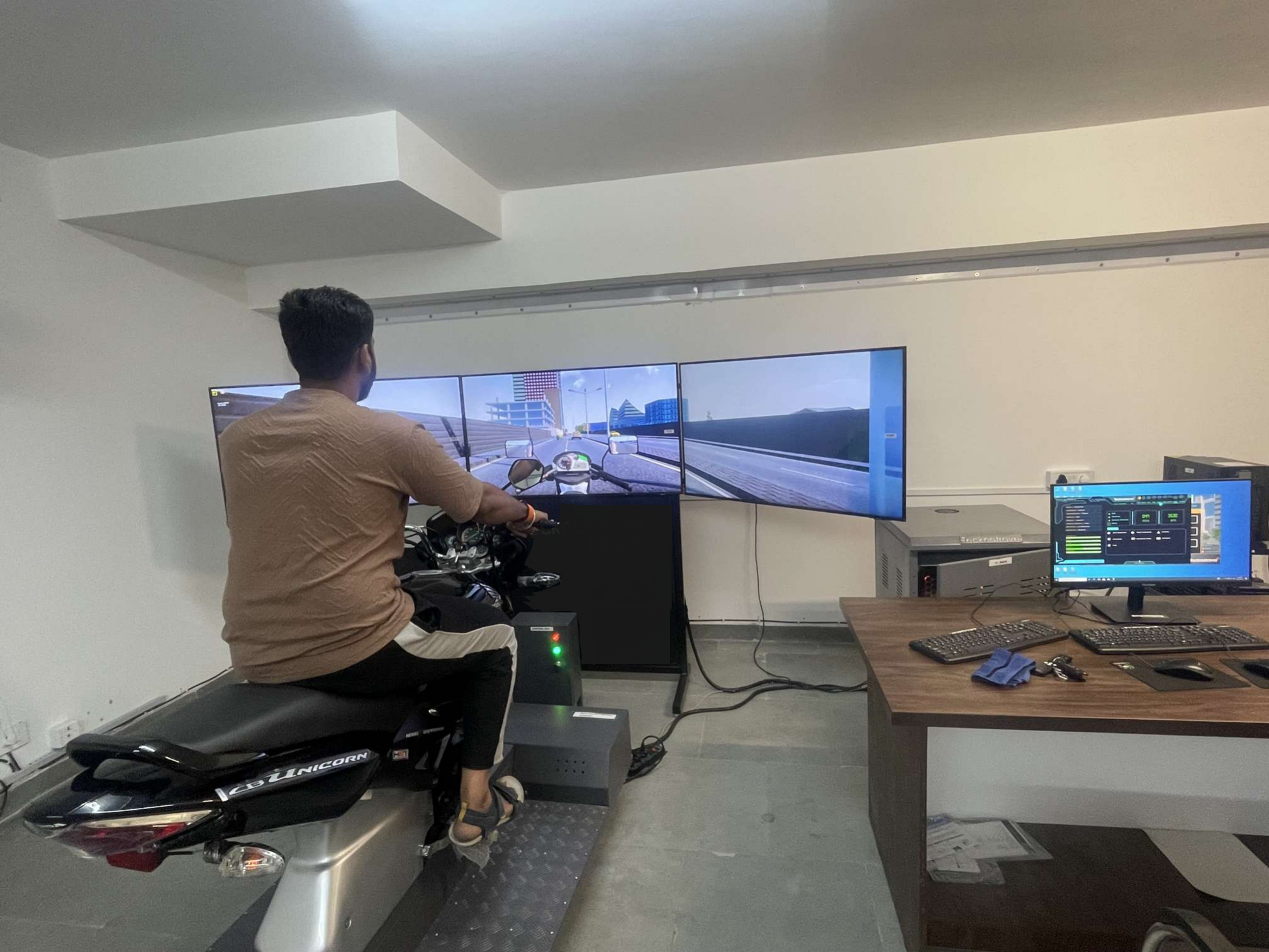}
        \label{fig:p2}
    \end{subfigure}
        \vspace{-0.2cm}
    \caption{Participants performing the simulator test.}
    \label{fig:participants_performing_test}
\end{figure}

We carried out all experimental trials on a fixed-base PTW simulator at our institute (Fig.~\ref{fig:participants_performing_test}). The simulator features a Honda motorcycle frame fitted with working throttle, brake, clutch, and gear controls, designed to replicate the riding experience of a real motorcycle. Riders are surrounded by a three-screen setup that delivers a wide viewing angle and lifelike road imagery. High-precision industrial-grade sensors continuously record rider inputs, vehicle dynamics, and environmental parameters through a real-time data acquisition interface, complying with IEC 393 accuracy standards~\citep{iec393}.  While the frame is static, a 4-actuator motion system provides vibration and acceleration.  Complete technical specifications, including sensor accuracy ratings, appear in Table~\ref{tab:simulator_specs}. This configuration offers a realistic yet controlled setting, allowing us to systematically study how riders behave across various traffic and road conditions.
\begin{table}
\centering
\scriptsize
\caption{Two-wheeler simulator: key technical parameters}
\label{tab:simulator_specs}
\renewcommand{\arraystretch}{1} 
\begin{tabular}{@{}lp{0.79\linewidth}@{}}
\toprule
\textbf{Component} & \textbf{Details} \\
\midrule
Platform & Fixed-base PTW frame with ISO certification \\
Controls & Acceleration, clutch, hydraulic brake unit, 5-speed shifting mechanism, handlebar steering, parking brake \\

Sensors & Wire-wound servo potentiometers (type 50WW): \\
& Resistance: 5 k$\Omega$ (±10\%); Linearity error: ±0.5\% (IEC 393 compliant), Angular range: 355° ± 3°; Full mechanical rotation: 360° \\
& Maximum power: 3 W (at 70°C); Expected operational life: 2 million cycles \\
& Safe operating limits: –40°C to +105°C \\
& Force transducers: Brake force sensing \\
& Rotary encoders: Gear position detection; SPDT microswitches: Gear shift sensing \\
Visual System & Three 50-inch LED panels with 180° horizontal coverage \\
Motion System & Four 3-DOF electric actuators capable of ±1 G max acceleration, 0–100 Hz response \\
& Motion ranges: Pitch ±1.75°, Roll ±3.5°, Heave 38.1 mm; Actuator stroke: 1.5 inches \\
& Per-actuator load rating: 114 kg \\
Computer Machine & Rider station: Intel i7, 32 GB RAM, GTX 1650 graphics card \\
& Operator station: Intel i5 processor, 16GB RAM, GTX 1650 GPU \\
Software & TechnoSim (AI-powered traffic, environment, and scenario simulation) \\
Data Logging & Continuous data recording with sampling frequency $f_s \geq 100$ Hz \\
\bottomrule
\end{tabular}
\end{table}

This ISO-compliant simulator provides a controlled, scalable, and safe environment to investigate rider behavior, cognitive load, and collision risk under realistic operational conditions.

\subsubsection{Simulated Scenarios}

\begin{figure}
    \centering
    \begin{subfigure}{0.48\textwidth}
        \centering
        \includegraphics[width=\textwidth]{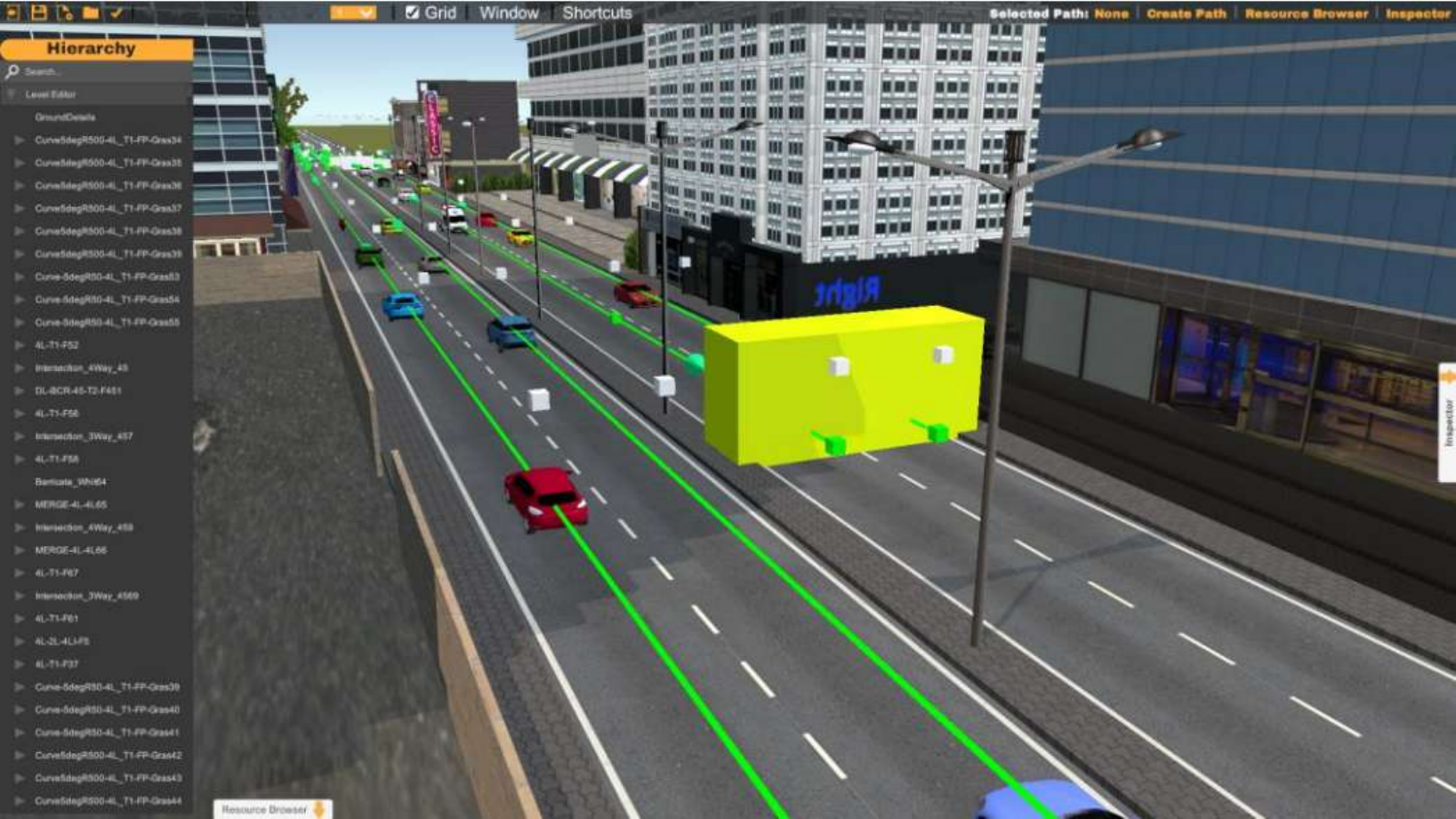}
        \subcaption{Primary route with intersections and conflict zones}
        \label{fig:simulated_scenarios:a}
    \end{subfigure}
    \hfill
    \begin{subfigure}{0.48\textwidth}
        \centering
        \includegraphics[width=\textwidth]{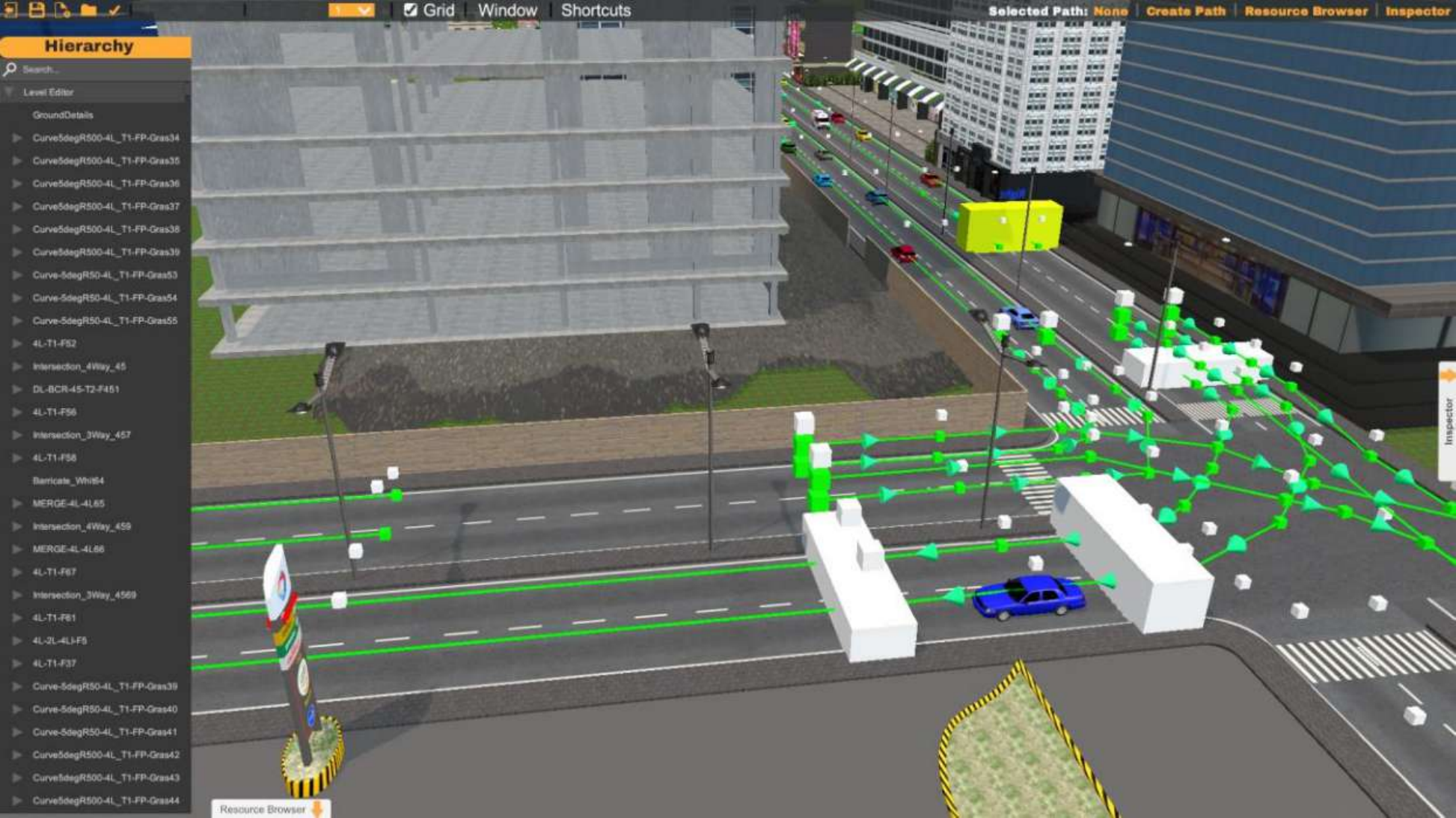}
        \subcaption{AI vehicle triggers simulating dynamic interactions}
        \label{fig:simulated_scenarios:b}
    \end{subfigure}
    
    \vspace{1em}
    
    \begin{subfigure}{0.48\textwidth}
        \centering
        \includegraphics[width=\textwidth]{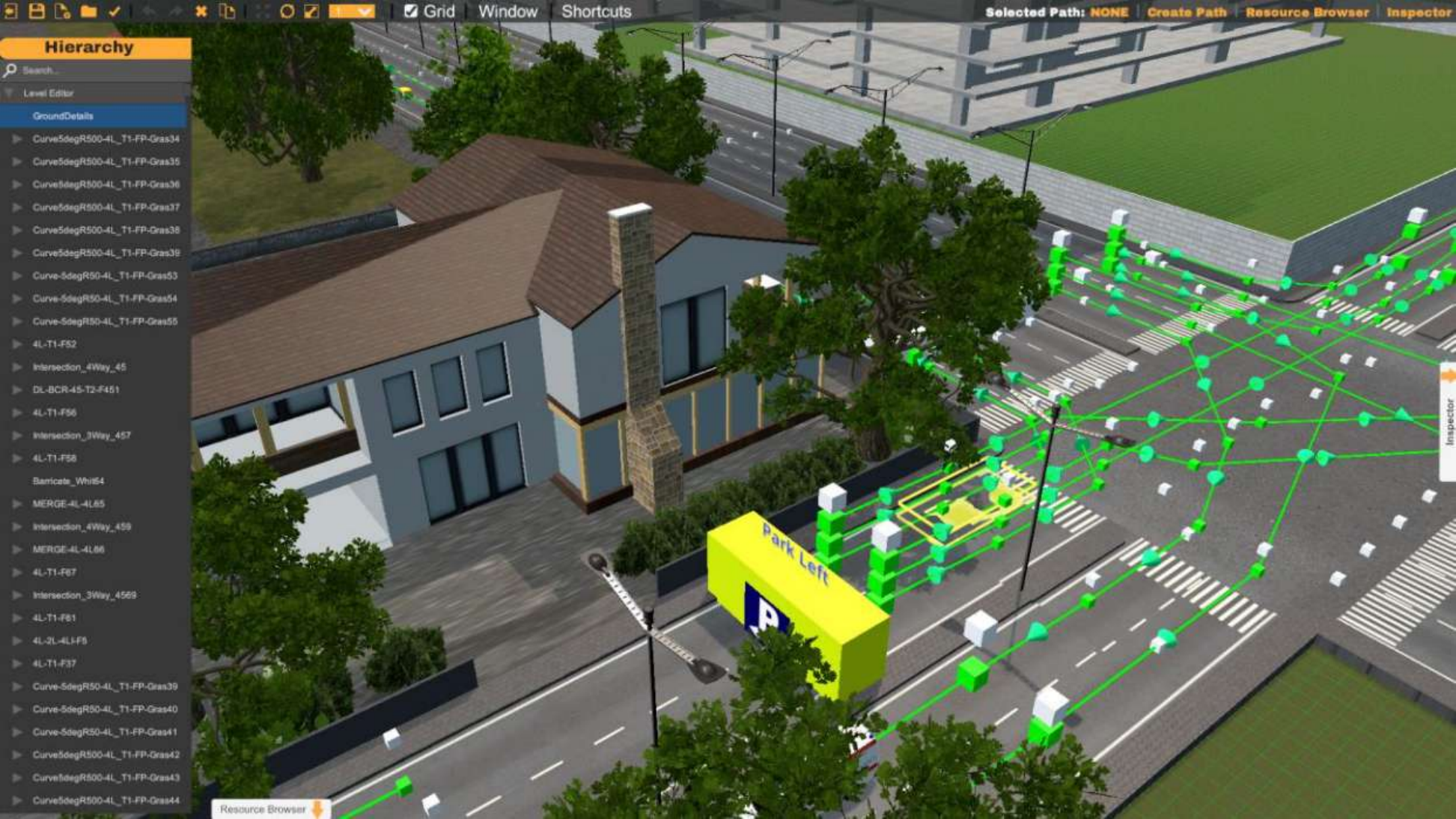}
        \subcaption{Route direction changes and detour prompts}
        \label{fig:simulated_scenarios:c}
    \end{subfigure}
    \hfill
    \begin{subfigure}{0.48\textwidth}
        \centering
        \includegraphics[width=\textwidth]{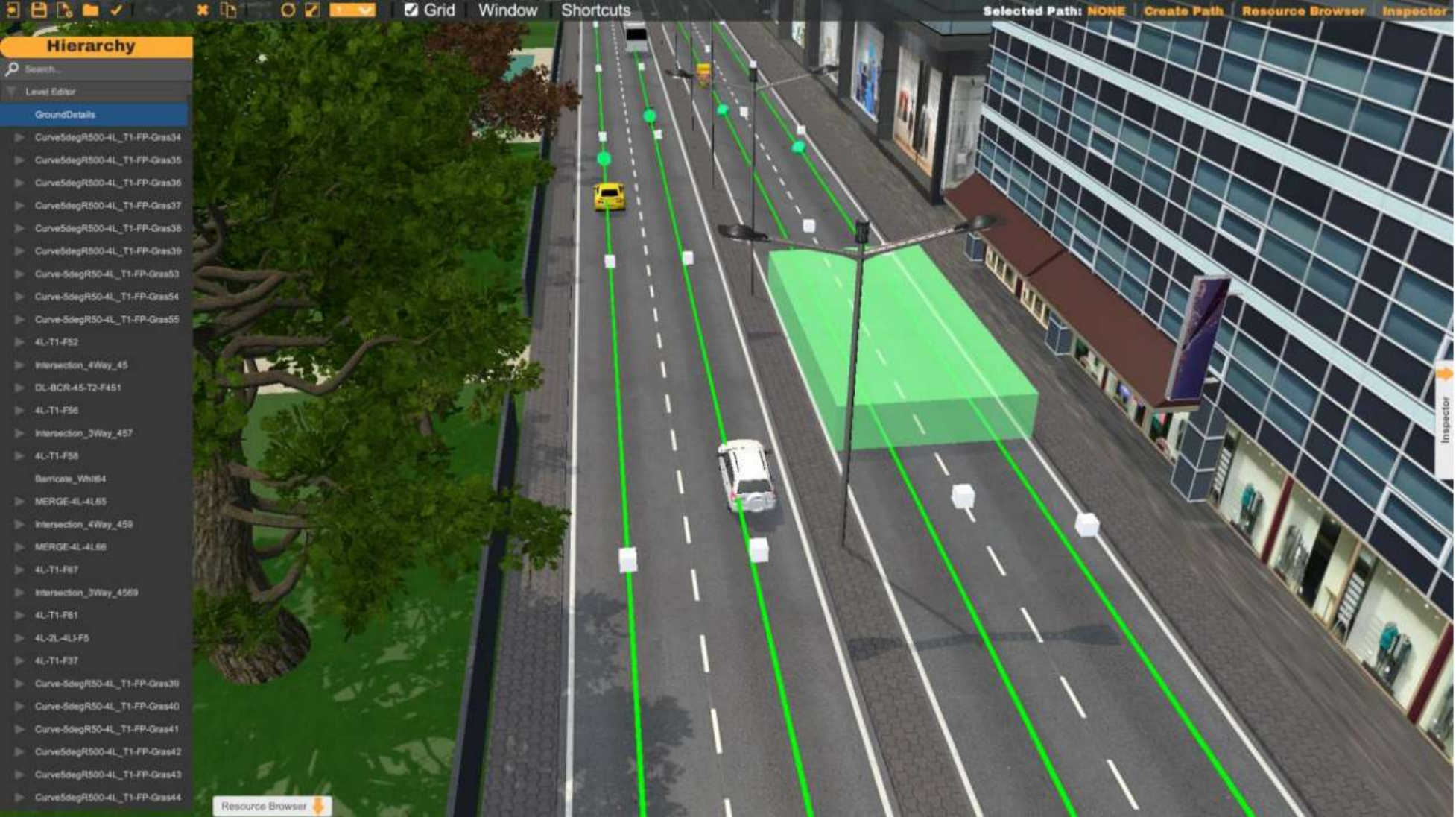}
        \subcaption{Speed change triggers near sensitive areas}
        \label{fig:simulated_scenarios:d}
    \end{subfigure}
    
    \caption{Sample snapshots of the simulated riding environment.}
    \label{fig:simulated_scenarios}
\end{figure}

We design a 4.8 km urban scenario on a high-fidelity two-wheeler simulator to investigate rider behavior under cognitive stress (TP). The route comprises a combination of two-lane undivided and four-lane road segments, with a maximum speed of 50~km/h, representing common road conditions found in Indian cities. The scenario embeds diverse events—including pedestrian crossings, obstacle overtaking, intersections, bike and car following segments to elicit real-time decision-making, adaptive control, and risk-taking under varying TP. The key elements of the scenario are illustrated in Fig.~\ref{fig:simulated_scenarios}: Fig.~\ref{fig:simulated_scenarios:a}: primary route with intersections and conflict zones, Fig.~\ref{fig:simulated_scenarios:b}: AI vehicle triggers simulating dynamic interactions, Fig.~\ref{fig:simulated_scenarios:c}: route direction changes and detour prompts, and Fig.~\ref{fig:simulated_scenarios:d}: speed change triggers near sensitive areas.

\subsubsection{Participants}
\label{subsec:participants}

\begin{table}[t]
\centering
\scriptsize
\caption{Demographic and riding profile of participants (N = 51)}
\label{tab:participant_stats}
\renewcommand{\arraystretch}{1}
\begin{tabular}{@{}lll@{}}
\toprule
\textbf{Attribute} & \textbf{Category / Range} & \textbf{Mean (SD) or \%} \\
\midrule
Age (years) & 18--42 & 26.4 (5.7) \\
Riding experience (years) & $\ge 2$ & 5.6 (3.1) \\
License status & Valid PTW license & 100\% \\
Education level & Graduate / B.Tech & 58.8\% / 41.2\% \\
Simulator exposure & No / Yes & 92.2\% / 7.8\% \\
Health status & Medically fit & 100\% \\
\bottomrule
\end{tabular}
\end{table}

MoRTH fatality data from 2019 to 2023 consistently shows that male riders constitute the overwhelming share of two-wheeler deaths in India, with yearly proportions of 86.0\%, 87.3\%, 86.4\%, 86.2\%, and 85.2\%, respectively~\citep{morth2019,morth2020,morth2021,morth2022,morth2023}. The 18–45 age group accounts for the largest share of these deaths. This strong demographic pattern supports our choice to include only male riders in this study. We therefore recruited 51 male riders aged 18 to 42 years, all with a minimum of two years of riding experience. Table~\ref{tab:participant_stats} summarizes their characteristics, while Fig.~\ref{fig:participants_performing_test} shows participants during simulator trials.

The male-only cohort is justified by three converging factors: First, national crash data shows male riders as the dominant high-risk group. According to MoRTH reports, male PTW fatalities accounted for 85.2--87.3\% of all PTW deaths from 2019 to 2023~\citep{morth2019,morth2020,morth2021,morth2022,morth2023}, with fatalities concentrated in the 18--45 age group. Statistical tests confirm this pattern: a binomial test shows male fatalities (86\%) significantly exceed equal representation ($p < 0.0001$), and a chi-square test confirms consistency across years ($\chi^2 = 262.3$, df = 5, $p < 0.0001$).

Second, female exposure to manually geared PTWs is extremely limited. Women hold only about 6\% of motor vehicle licenses in India~\citep{GUPTA2026108408,tripcentre2023roadsafety,atif2026modeling}, significantly below equal representation ($p < 0.0001$). The number of women riding manually geared motorcycles is even smaller, largely due to socio-cultural factors and traditional norms and practical barriers~\citep{tripcentre2023roadsafety}. Consequently, the pool of female riders with manual transmission experience suitable for simulator-based research is extremely small.

Third, the simulator's manual transmission requirement creates a practical recruitment constraint. The high-fidelity PTW simulator replicates a manually geared motorcycle with functional clutch and gear controls, requiring prior manual transmission experience. Given the extremely low prevalence of female riders with such experience in India, recruiting a statistically sufficient female sample for robust analysis is infeasible. Thus, male participants were selected to ensure a sufficiently large and statistically robust sample for meaningful spatial risk analysis, consistent with the dominant high-risk demographic. Future studies will include female riders as their PTW participation increases.

\subsubsection{Time Pressure (TP) Scenarios}
To ethically evoke graduated cognitive load mimicking emergency commuting, participants were tasked with reaching an examination hall under three conditions: (i) NTP: Baseline with ample time; (ii) LTP: Limited to 90\% of the baseline time (with a reminder to avoid being late); and (iii) HTP: Restricted to 80\% of the baseline time, accompanied by active urgency cues (e.g., exam hall closing shortly). This standardized exam-deadline scenario simulates high-arousal stress states common in urban transit, providing a controlled environment to study behavioral degradation.

\subsubsection{Experimental Design and Protocol}
\label{subsec:protocol}

\begin{figure}
    \centering
    \includegraphics[width=0.83\textwidth]{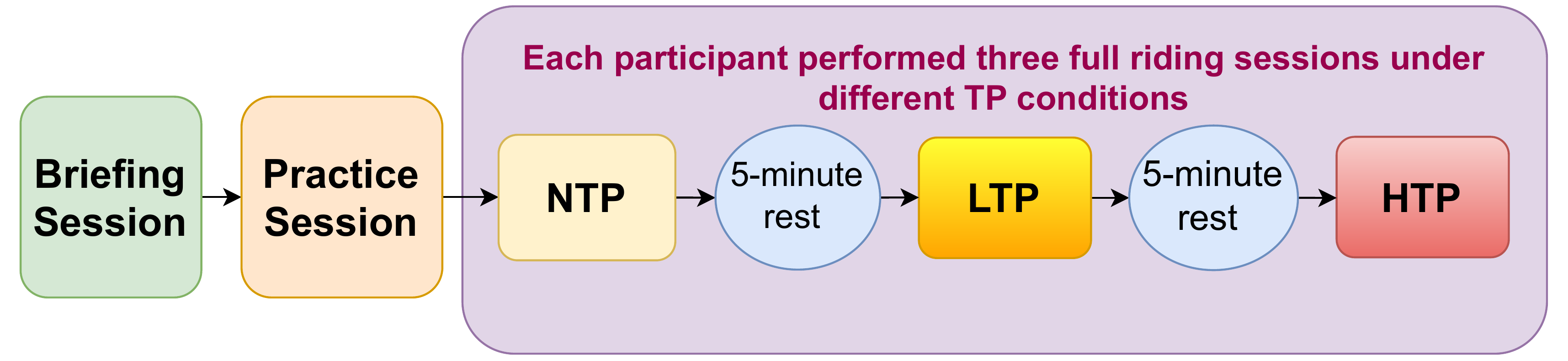}
    \caption{Flow of dataset collection and experimental design.}
    \label{fig:experiment_flow}
\end{figure}

A structured four-phase experimental protocol (Fig.~\ref{fig:experiment_flow}) was implemented to ensure data consistency: (i) Briefing: Standardized orientation and informed consent; (ii) Practice: A 5--10 minute familiarization ride (excluded from analysis); (iii) Main Task: Participants completed three riding sessions under NTP, LTP, and HTP conditions, with the session sequence randomized across participants to avoid order effects; and (iv) Rest: 5--8 minute inter-session breaks to stabilize cognitive load and mitigate fatigue.

\textbf{Ethics and Informed Consent:} All participants signed a written consent form before taking part in the research. The research protocol received ethical clearance from the IE Committee at the IIT Indore (Approval No. BSBE/IITI/IHEC-11/2025/11).

\textbf{Consent to Publish:} All participants gave their written consent for the publication of anonymized data and findings derived from this study. To safeguard participant privacy, all data have been fully anonymized.

\subsubsection{Behavioral Transition : Why LTP Matters}

Adding the LTP condition allows us to capture the transitional phase from safe riding (NTP) to risky riding (HTP), representing the early stages of behavioral decline. The progression \textit{NTP $\rightarrow$ LTP $\rightarrow$ HTP} illustrates the transition from stable, safe riding to high-stress operation characterized by speeding and infractions. LTP serves as a critical indicator where subtle warning signs—such as minor speeding or slight lane drifts—first manifest. By incorporating LTP, the model improves its sensitivity to rising stress levels, supporting earlier detection and enabling more timely safety interventions. Moreover, LTP data inform policymakers and trainers by defining practical stress thresholds, facilitating preventive strategies before riders reach the high-risk HTP state.

\subsection{Feature Engineering and Preprocessing}

From raw time-series sensor data, we derived 64 domain-informed features that capture vehicle dynamics, rider behavior, and contextual factors as summarized in Table~\ref{tab:feature_summary}. The features include things like average and variation in speed and acceleration, how often the rider brakes, how frequently they change lanes, and the TP level. Min-Max normalization was used to bring all feature values within the $[0,1]$ range. We segment the time series data into fixed-length overlapping windows to preserve temporal dependencies, enabling the model to capture both gradual risk accumulation and abrupt behavioral deviations. Detailed training configurations are provided in Section~\ref{sec:training}.

\subsection{Dataset and Feature Set}
\label{sec:Dataset_Details}

The experimental protocol yielded 153 multivariate time-series sessions (51 participants $\times$ 3 conditions). Data were sampled at 100~Hz. The 100 Hz streams were segmented into 960 ms windows (96 time steps) with 50\% overlap, yielding 129,209 windows slid continuously. A window is labeled as collision (1) if a collision occurs within it, otherwise 0. The dataset exhibits a class distribution of 21\% collision and 79\% non-collision events. A single collision spans multiple windows (impact+sliding+aftermath, due to overlap), reflecting the inherent imbalance of safety-critical scenarios. This is window-level, not event-level labeling. 

Each input sequence comprises 64 features (summarized in Table~\ref{tab:feature_summary}) categorized into: (i) Vehicle Dynamics: Speed, acceleration, and 3D rotation; (ii) Control Inputs: Throttle, hydraulic brake force, and gear transitions; (iii) Proximity: Headway/tailway distances and lane offsets; (iv) Time context and scenario: Time Stamp, TP (0=HTP, 1=LTP, 2=NTP); and (v) Behavioral Violations: Overspeeding, improper gap maintenance and traffic rule infractions. All collisions were aggregated into a single binary target variable, with leakage prevented by excluding impact-related features from the input.

\begin{table*}
\centering
\caption{Overview of simulator-derived features and their categories}
\label{tab:feature_summary}
\renewcommand{\arraystretch}{1} 
\setlength{\tabcolsep}{8pt}
\scriptsize
\begin{tabular}{@{}p{3.5cm} p{12cm} c@{}}
\toprule
\textbf{Feature Category} & \textbf{Included Parameters} & \textbf{Count} \\
\midrule
\textbf{Vehicle Controls} & Ignition status, Engine state, Throttle position, Brake application, Clutch engagement, Handbrake, Steering angle, Gear position, Headlight usage, Horn activation & 10 \\
\textbf{Vehicle Performance} & Speed, Engine RPM, Fuel consumption, Distance covered & 4 \\
\textbf{Lighting \& Indicators} & Indicator signal, Pre-move indication, Junction turning signal, Lane-change signal, Headlight non-compliance & 5 \\
\textbf{Behavioral Violations} & Speeding, Incorrect junction/intersection speed, Speed violations on speed breakers, Gap maintenance errors, Hazardous overtaking, Unsigned turns, Lane misuse, Driving on the wrong side, Handbrake use while moving, Riding the clutch, Faulty gear shift order, Improper clutch release, Shifting without clutch engagement, Proper gear selection before start, Gradual clutch engagement & 15 \\
\textbf{Traffic Infractions} & White line violations, Yellow line crossings, Stop line breaches, Signal jumping, Entering no-entry zones, Illegal U-turns, Parking infringements & 7 \\
\textbf{Temporal Context} & Time stamp, Time Pressure condition (0=HTP, 1=LTP, 2=NTP) & 2 \\
\textbf{Spatial Attributes} & 3D position coordinates, 3D rotation angles, Lane number, Left/Right lane offsets & 9 \\
\textbf{Motion and Proximity} & Lateral/Longitudinal velocity, Headway (distance \& time), Tailway (distance \& time), Left/Rightway distance, Steering angle & 9 \\
\textbf{Brake System} & Brake test completion, Front and Rear braking force & 3 \\
\bottomrule
\end{tabular}
\\[2pt]
\footnotesize{\textit{Note: All features associated with collisions (vehicles, objects, obstacles, etc.) were merged into a single binary target variable `Target`, where 0 = no collision and 1 = collision. These collision events were excluded from the input feature set to prevent label leakage.}}
\end{table*}

\subsection{Scope of the Classification and Forecasting Tasks}

We define two complementary but distinct tasks, and are explicit about what each does and does not establish.

\textbf{Classification Task (Same-Window Risk Detection):} Given a 960~ms observation window, the model predicts whether a collision event occurs within that window. Because a subset of positive windows overlaps the collision event itself (impact, sliding, or aftermath phases, as noted in Section~\ref{sec:Dataset_Details}), this task is best characterized as \emph{same-window collision risk detection} rather than a fixed-lead-time forecast: it evaluates whether pre-collision and in-progress behavioral signatures are jointly separable from normal riding, not how far in advance a collision can be anticipated.

\textbf{Forecasting Task (Prospective Kinematic Prediction):} Given a lookback window of $L=96$ timesteps, a separately trained instance of the architecture predicts future \emph{kinematic states} (e.g., speed, lean angle, longitudinal force) up to $H \in \{96, 192, 336, 720\}$ timesteps (up to 7.2s) ahead. This task, evaluated in Section~\ref{sec:forecasting_results}, is a genuine prospective-forecasting result with an explicit lead time, but its target is future sensor state, not future collision probability; it demonstrates that the architecture can anticipate the kinematic precursors of hazardous states, which is complementary to, but distinct from, a direct lead-time collision-probability forecast.

We view a labeled, fixed-lead-time collision-probability forecast (i.e., excluding all windows that overlap a collision event and labeling remaining windows by whether a collision begins within a defined horizon $\tau$ afterward) as an important direction for follow-up work, and report the current classification results with this scope explicitly stated rather than implied.

\subsection{MotoSafety: Proposed Architecture}

\subsubsection{Architecture Overview}

\begin{figure}[t]
\centering
\includegraphics[width=0.75\textwidth]{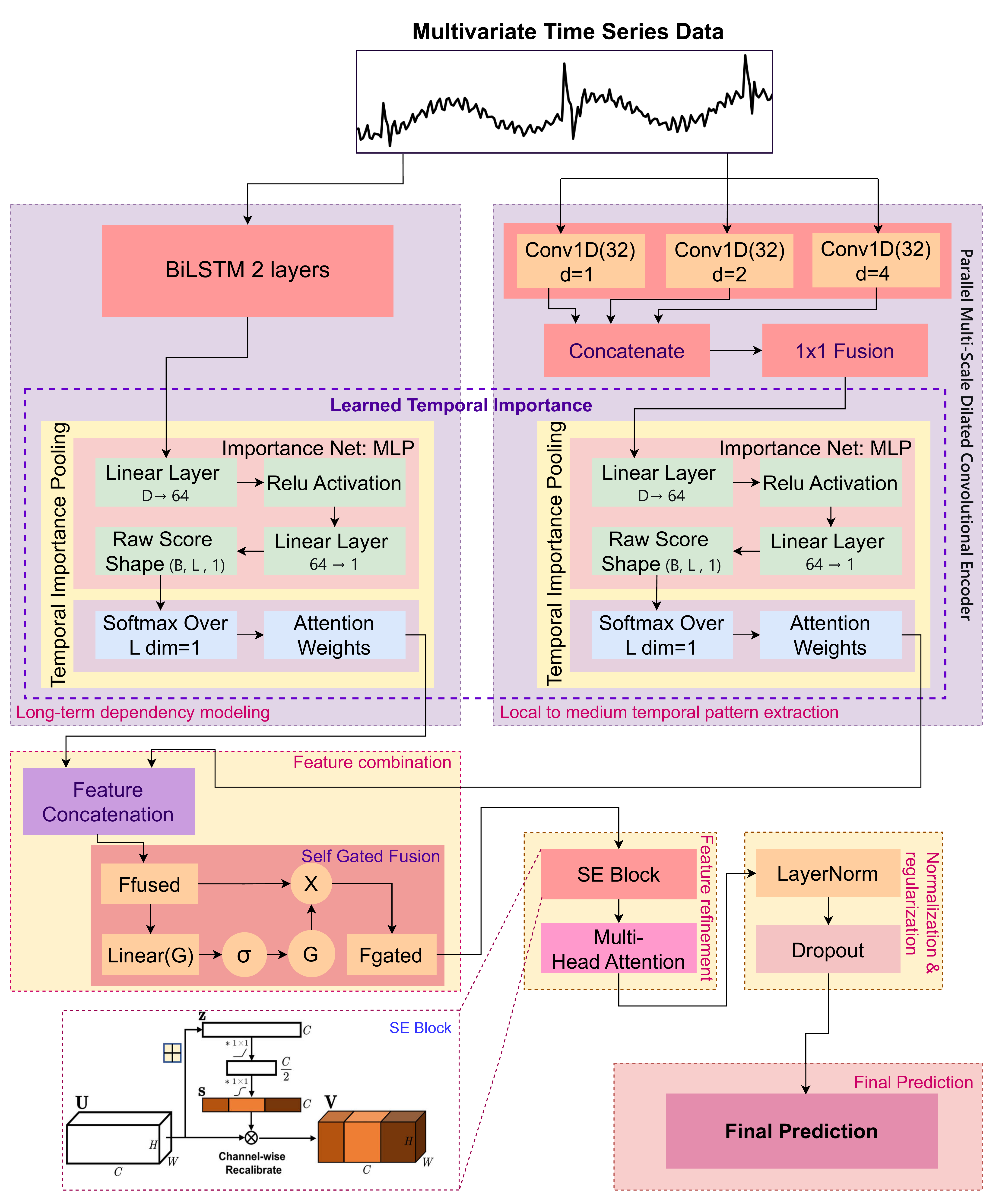}
\caption{The proposed MotoSafety architecture.}
\label{fig:MST_architecture}
\end{figure}

The proposed MotoSafety architecture, shown in Fig.~\ref{fig:MST_architecture}, is a DL architecture based on the Learned Temporal Importance (LTI) principle, developed for the prediction of PTW collisions and forecasting. It employs a multi-stage pipeline: (i) a parallel multi-scale dilated convolutional encoder with three branches (dilation factors 1, 2, 4) extracts local to medium-range temporal patterns; (ii) a 2-layer Bi-LSTM (128 hidden units per direction) captures long-range sequential dependencies; (iii) a novel TIP module (which uses a lightweight Multi-Layer Perceptron (MLP) as a learned scoring function to evaluate the importance of each timestep) is applied independently to the outputs of both the CNN and LSTM paths, learning to adaptively weight the most critical pre-collision timesteps for each feature type and collapsing them into compact vectors; (iv) these vectors are concatenated and processed by a self-gated fusion stage, refined by channel-wise recalibration via a SE block and MHA (4 heads). 
This representation undergoes normalization, dropout-based regularization, and linear projection to output binary collision risk scores. 


\subsubsection{Comment on the Novelty}

We distinguish clearly between architectural components adapted from prior work and the components that constitute this paper's novel contribution.

\textbf{Adapted components:} The parallel dilated-convolution encoder, Bi-LSTM, SE recalibration, and MHA are established building blocks, each individually well-studied in time-series and sequence modeling literature.

\textbf{Novel contribution: TIP.} The core contribution of this work is the TIP module and the LTI principle it instantiates. Rather than collapsing the temporal dimension with a fixed operator (mean-pool, max-pool, or final-hidden-state extraction, as is standard practice in CNN/RNN-based time-series classifiers), TIP learns a content-aware scoring function that assigns a data-dependent importance weight to every timestep before collapsing the sequence. This is applied independently to both the CNN and BiLSTM representations, allowing the model to learn which pre-collision moments matter most for each feature pathway, rather than treating all timesteps as equally informative. Unlike full self-attention pooling (which computes pairwise timestep interactions at $\mathcal{O}(L^2)$ cost), TIP scores each timestep independently against a shared learned criterion, at $\mathcal{O}(L)$ cost, and produces a fixed-size vector consumed by the downstream fusion stage. Applied independently to both CNN and BiLSTM branches, TIP reduces the downstream fusion, SE-recalibration, and attention stages to $\mathcal{O}(1)$ relative to the sequence length, while the front-end encoder operates at $\mathcal{O}(L)$ complexity, resulting in an overall $\mathcal{O}(L)$ framework (Table~\ref{tab:complexity}).

\textbf{Empirical validation of TIP.} The effectiveness of TIP is validated in Section~\ref{sec:Ablation} (Table~\ref{tab:tip_ablation}), where it outperforms standard fixed-pooling strategies (mean-pooling, max-pooling, last-timestep) under an identical architecture, indicating that the performance gain stems from learned, content-aware weighting rather than fixed pooling.

This strategic ``content-aware collapse'' means the fusion, SE-recalibration, and MHA stages operate on a fixed-size 320-dimensional vector regardless of sequence length, i.e., these downstream stages cost $\mathcal{O}(1)$ per sample once pooling is complete. The upstream CNN and Bi-LSTM encoder remains $\mathcal{O}(L)$, so the model's end-to-end complexity is $\mathcal{O}(L)$ overall, not $\mathcal{O}(1)$; the practical benefit of TIP is that it removes the $\mathcal{O}(L^2)$ cost a full-sequence self-attention mechanism would otherwise add on top of the encoder. Table~\ref{tab:complexity} details the per-module accounting. This design ensures the low-latency performance required for robust real-time deployment on edge devices.

\begin{table}
\centering
\scriptsize
\caption{Per-module computational complexity. $L$: sequence length, $C$: feature channels, $H$: hidden dimension, $k$: kernel size.}
\label{tab:complexity}
\begin{tabular}{ll}
\toprule
\textbf{Module} & \textbf{Complexity} \\
\midrule
Dilated CNN encoder (×3 branches) & $\mathcal{O}(L \cdot k \cdot C)$ \\
Bi-LSTM (2 layers) & $\mathcal{O}(L \cdot H^2)$ \\
TIP scoring + weighted sum & $\mathcal{O}(L \cdot H)$ \\
Gated fusion + SE block & $\mathcal{O}(1)$ (fixed-size input) \\
Multi-head attention (post-pooling) & $\mathcal{O}(1)$ (fixed-size input) \\
\midrule
\textbf{MotoSafety total} & $\mathcal{O}(L)$ \\
\bottomrule
\end{tabular}
\end{table}

\subsubsection{MotoSafety: Mathematical Formulation}
\label{sec:math_model}

We address the problems of binary collision prediction and multi-horizon time-series forecasting from multivariate sensor data under TP. Our input consists of sensor readings $X \in \mathbb{R}^{B \times L \times C}$ from a two-wheeler simulator, with $B$ batch size, lookback window $L$ and $C$ feature channels. 

\textbf{Forward Process:} The input $X$ is processed by two parallel branches: CNN and BiLSTM.  The CNN branch captures local patterns using three dilated convolutional branches with kernel size $k=3$. To accommodate the channel-first requirement of 1D convolutions, we apply a permutation function $\pi(\cdot)$ to $X$ such that $\hat{X} = \pi(X) \in \mathbb{R}^{B \times C \times L}$:
\begin{align}
    H_{\text{d}} &= \text{ReLU}(\text{Conv1D}_{d}(\hat{X}))
\end{align}
where $d =\{1,2,4\}$ is the dilation factor. These views are merged into a unified 64-channel representation via a $1\times1$ convolution: $H_{\text{ms}} = \text{Conv1D}_{1\times1}([H_{\text{1}}; H_{\text{2}}; H_{\text{4}}]) \in \mathbb{R}^{B \times 64 \times L}$. In parallel, a bidirectional LSTM captures long-range dynamics:
\begin{equation}
    H_{\text{lstm}} = \text{BiLSTM}(X) \in \mathbb{R}^{B \times L \times 256}.
\end{equation}

To identify safety-critical moments, we utilize TIP to learn importance weights across the temporal dimension. Two distinct networks, $\text{MLP}_{\theta_1}$ and $\text{MLP}_{\theta_2}$, assign scores to the CNN and LSTM features respectively:
\begin{small}
\begin{align}
    w_{\text{c}} &= \text{Softmax}(\text{MLP}_{\theta_1}(H_{\text{ms}}^T)), \quad f_{\text{cnn}} = \sum_{t=1}^{L} w_{\text{c}}^{(t)} H_{\text{ms}}^{(t)} \\
    w_{\text{l}} &= \text{Softmax}(\text{MLP}_{\theta_2}(H_{\text{lstm}})), \quad f_{\text{lstm}} = \sum_{t=1}^{L} w_{\text{l}}^{(t)} H_{\text{lstm}}^{(t)}
\end{align}
\end{small}
The resulting vectors, $f_{\text{cnn}} \in \mathbb{R}^{64}$ and $f_{\text{lstm}} \in \mathbb{R}^{256}$, are concatenated to form the joint representation $f = [f_{\text{cnn}}, f_{\text{lstm}}] \in \mathbb{R}^{320}$. The fused representation undergoes refinement through a gating mechanism and a channel-wise recalibration block (SE-Block). Finally, a MHA block performs high-order feature refinement:
\begin{align}
    f_{\text{gate}} &= f \odot \sigma(W_g f + b_g), \\
    f_{\text{se}} &= \text{SEBlock}(f_{\text{gate}}), \\
    f_{\text{attn}} &= \text{MHA}(f_{\text{se}}).
\end{align}
The final output $\mathbf{y} \in \mathbb{R}^2$ is produced by applying a linear layer to the normalized and regularized context representation:
\begin{equation}
    \mathbf{y} = \text{Softmax}\Bigl(\text{Linear}\bigl(\text{LayerNorm}(\text{Dropout}(f_{\text{attn}})\bigr)\Bigr).
\end{equation}

To address the extreme class imbalance (rare collision events), we utilize Focal Loss with a focusing parameter $\gamma=2$:
\begin{equation}
    \mathcal{L}_{\text{cls}} = -\frac{1}{B} \sum_{i=1}^{B} (1 - P_{i, y_i})^{\gamma} \log(P_{i, y_i}),
\end{equation}
where $P_{i, y_i}$ denotes the model's estimated probability for the actual binary collision outcome $y_i \in \{0,1\}$.

\paragraph{Multi-Horizon Forecasting Module}
Our forecasting module predicts future target values $\hat{Y} \in \mathbb{R}^{B \times \tau}$ over horizon $\tau$. It uses a \textit{separate but architecturally similar} network with identical CNN and BiLSTM branches, TIP, gating, SE-Block, and MHA. Both the classification and forecasting modules use the same $L=96$ timestep lookback window; they differ only in their prediction target (same-window collision label vs.\ future kinematic states) and are trained separately, as described above. This produces a refined context vector $\mathbf{f}_{\text{attn}} \in \mathbb{R}^{B \times 320}$. The final forecast is obtained via:
\begin{equation}
    \hat{Y} = \text{Linear}_{\text{forecast}}(\text{LayerNorm}(\mathbf{f}_{\text{attn}})).
\end{equation}

The module is trained to minimize the Mean Squared Error (MSE) loss:
\begin{equation}
\mathcal{L}_{\text{forecast}} = \frac{1}{B \cdot \tau} \sum_{i=1}^{B} \sum_{k=1}^{\tau} \left( y_{i, L+k} - \hat{Y}_{i, k} \right)^2,
\end{equation}
where $y_{i, L+k}$ is the true target value at future step $k$. \textit{Both models share identical architectural components but are trained separately for their respective tasks.}

\section{Experimental Setup and Baselines}

\subsubsection{Ground Truth and Human Validation}

To ensure the reliability of the simulator-generated labels, a systematic validation protocol was conducted. Two road experts in road safety and human factors (male, post-graduate) reviewed a representative subset of 2,000 randomly selected segments from the total of 129,209 feature windows. Experts independently annotated the onset and severity of high-risk episodes (e.g. abrupt swerves, loss of balance, critical near-misses) using a standardized coding scheme based on kinematic thresholds and behavioral cues. We used Cohen's Kappa ($\kappa$) to check how well the two annotators agreed with each other, which yielded a value of 0.87. This indicates near-perfect agreement based on established interpretation standards~\citep{landis1977measurement}. This high agreement validates the dataset's labels as a robust ground truth for model training.

\subsubsection{Baselines}

We benchmark MotoSafety against ML (RF, CNN, RNN) and modern DL architectures: Informer \citep{zhou2021informer} and iTransformer \citep{liu2024itransformerinvertedtransformerseffective} for Transformer-based long-range modeling. TimesNet \citep{wu2023timesnettemporal2dvariationmodeling} for multi-scale frequency analysis; PatchTST \citep{nie2023timeseriesworth64} for patching-based local semantics. Time-LLM \citep{jin2024timellmtimeseriesforecasting} and LLM4TS (GPT-2) \citep{10.1145/3719207} as a representative fine-tuned LLM baseline.

\subsection{Training and Evaluation}
\label{sec:training}

\subsubsection{Training Protocol}

The MotoSafety model is implemented in PyTorch and optimized using AdamW (initial learning rate: $3\times10^{-4}$, L2 weight decay: $5\times10^{-5}$) with Focal Loss ($\gamma=2$) for classification and MSE loss for forecasting. To enhance generalization, we employed Mixup augmentation, Exponential Moving Average (EMA) of weights, and Monte Carlo (MC) Dropout (5 samples) for uncertainty calibration. For model training, we utilized an NVIDIA T400 GPU (4 GB VRAM) and set the batch size to 64, running the process for 50 epochs. The main training settings are listed in Table~\ref{tab:training_config}. For all experiments, we used a participant-wise data split of 80\% for training, 10\% for validation, and 10\% for testing, where all sessions from the same rider are kept together in the same split to prevent data leakage.

\begin{table}[t]
\centering
\scriptsize
\caption{Training configuration for MotoSafety classification and forecasting models.}
\label{tab:training_config}
\begin{tabular}{ll}
\toprule
\textbf{Component} & \textbf{Setting} \\
\midrule
\multicolumn{2}{l}{\textbf{Classification Model}} \\
\midrule
Input sequence length & 96 timesteps (960 ms) \\
Feature dimension & 64 \\
CNN hidden channels & 32 per branch, 64 fused \\
BiLSTM hidden dimensions & 128 (bidirectional) \\
Total model parameters & $\sim$1.15 million \\
Dropout rate & 0.3 \\
Loss function & Focal Loss ($\gamma=2$) \\
\midrule
\multicolumn{2}{l}{\textbf{Forecasting Model}} \\
\midrule
Input lookback length & 96 timesteps \\
Prediction horizons & $H \in \{96, 192, 336, 720\}$ \\
Architecture & Identical to classification model \\
Loss function & MSE \\
\midrule
\multicolumn{2}{l}{\textbf{Shared Settings}} \\
\midrule
Optimizer & AdamW (initial learning rate: $3\times10^{-4}$, L2 weight decay: $5\times10^{-5}$) \\
Batch size & 64 \\
Training epochs & 50 \\
Regularization & EMA, Mixup, MC Dropout (5 samples) \\
Hardware & NVIDIA T400 GPU (4GB VRAM) \\
\bottomrule
\end{tabular}
\end{table}

\subsubsection{Evaluation Protocol}

We evaluated classification performance using Accuracy, F1-Score, and ROC-AUC, while for the forecasting task, we used MSE and Mean Absolute Error (MAE) as evaluation metrics. Statistical significance was assessed via paired Wilcoxon signed-rank tests over 5 runs, with the Holm--Bonferroni correction ($\alpha = 0.05$).

\section{Results}

\subsection{Impact of Time Pressure on Riding Behavior}

\begin{table}
\centering
\scriptsize
\caption{Behavioral differences across TP: NTP/LTP/HTP (Mean $\pm$ Std).}
\label{tab:tp_behavior}
\setlength{\tabcolsep}{3pt}
\begin{tabular}{llll}
\toprule
\textbf{Feature} & \textbf{NTP} & \textbf{LTP} & \textbf{HTP} \\
\midrule
Speed (km/h) & 33.12 $\pm$ 19.42 & 39.69 $\pm$ 21.67 & \textbf{49.00 $\pm$ 26.48} \\
Sudden braking & 1.32 $\pm$ 2.54 & 1.38 $\pm$ 2.36 & \textbf{1.80 $\pm$ 2.01} \\
Dangerous overtaking & 1.07 $\pm$ 1.18 & 1.21 $\pm$ 1.27 & \textbf{1.36 $\pm$ 1.52} \\
Headway distance (m) & 8.99 $\pm$ 14.38 & 8.20 $\pm$ 14.34 & \textbf{7.75 $\pm$ 13.88} \\
\bottomrule
\end{tabular}
\end{table}

Analysis of our PTW dataset reveals that HTP is the primary driver of behavioral volatility, with a clear transition to high-risk strategies as TP increases, as shown in Table~\ref{tab:tp_behavior}. Riders under HTP exhibited 48\% higher mean speed, 36\% increase in sudden braking, and 14\% shorter headway distance compared to NTP, along with a 27\% increase in dangerous overtaking. These statistically supported trends define a hurry-up strategy of excessive speed, abrupt control, and reduced spacing, highlighting TP as a key behavioral stressor that increases the risk of PTW collisions.

\begin{table}
\centering
\scriptsize
\caption{Multivariate long-term forecasting results (mean $\pm$ std over 5 runs). Input lookback length $L=96$. Forecast horizons $H \in \{96, 192, 336, 720\}$. Avg is averaged over all four prediction horizons.}
\label{tab:forecasting_results}
\setlength{\tabcolsep}{2pt} 
\begin{tabular}{l|cc|cc|cc|cc}
\toprule
\textbf{Models} & \multicolumn{2}{c|}{\textbf{MotoSafety}} & \multicolumn{2}{c|}{\textbf{Time-LLM}} & \multicolumn{2}{c|}{\textbf{iTransformer}} & \multicolumn{2}{c}{\textbf{PatchTST}} \\
\textbf{Metric} & \textbf{MSE} & \textbf{MAE} & \textbf{MSE} & \textbf{MAE} & \textbf{MSE} & \textbf{MAE} & \textbf{MSE} & \textbf{MAE} \\
\midrule
96  & \textbf{0.033 $\pm$0.005} & \textbf{0.082 $\pm$0.007} & 0.136 & 0.148 & 0.161 & 0.237 & 0.188 & 0.208 \\
192 & \textbf{0.037 $\pm$0.006} & \textbf{0.086 $\pm$0.008} & 0.145 & 0.153 & 0.170 & 0.246 & 0.205 & 0.223 \\
336 & \textbf{0.041 $\pm$0.006} & \textbf{0.097 $\pm$0.010} & 0.192 & 0.215 & 0.177 & 0.258 & 0.227 & 0.272 \\
720 & \textbf{0.045 $\pm$0.007} & \textbf{0.111 $\pm$0.012} & 0.212 & 0.293 & 0.183 & 0.262 & 0.267 & 0.338 \\
\midrule
\textbf{Avg} & \textbf{0.039} & \textbf{0.094} & 0.171 & 0.210 & 0.173 & 0.251 & 0.222 & 0.260 \\
\bottomrule
\end{tabular}
\end{table}

\subsection{Multivariate Long-Term Forecasting Performance}
\label{sec:forecasting_results}

To evaluate the proactive safety capabilities of MotoSafety, we conducted multivariate forecasting across four different time horizons ($H \in \{96, 192, 336, 720\}$) on our PTW data. This experiment assesses the models ability to project future kinematic states such as lean angles and longitudinal forces, essential for anticipating hazardous transitions before they culminate in a collision. As shown in Table~\ref{tab:forecasting_results}, MotoSafety attains a mean MSE of 0.039 and an MAE of 0.094, representing a 4.4$\times$ reduction in error over the Time-LLM (0.171) and iTransformer (0.173) baselines. This high forecasting fidelity under TP is critical for proactive safety; by accurately projecting states up to $H=720$, the model provides a vital safety buffer for early risk mitigation before a maneuver reaches a non-recoverable limit.

\subsection{Comparative Classification Performance}

    The model performance results on the PTW simulation dataset under TP are presented in Table~\ref{tab:model_performance}. The proposed MotoSafety achieves 94.97\% accuracy, 93.7\% F1-score and 99.33\% ROC AUC, outperforming ten baselines including TimesNet, PatchTST, iTransformer, Time-LLM and LLM4TS. To assess the robustness of the performance gains, statistical significance is evaluated using paired Wilcoxon signed-rank tests over 5 independent runs (Table~\ref{tab:model_performance}). For the nine pairwise comparisons against baselines, we applied the Holm--Bonferroni method ($\alpha = 0.05$) to regulate the family-wise error rate (FWER). The high ROC AUC indicates that MotoSafety can maintain a low false-alarm rate, which is a prerequisite for real-world Advanced Rider Assistance Systems (ARAS).

\begin{table}
\centering
\scriptsize
\begin{minipage}{\linewidth} 
\centering
  \caption{Comparison of model performance (mean $\pm$ standard deviation over 5 runs). We used the paired Wilcoxon signed-rank test with Holm--Bonferroni adjustment to determine statistical significance ($\alpha = 0.05$). Significance is denoted by: $^{***}p < 0.001$, $^{**}p < 0.01$, $^{*}p < 0.05$. All baseline models were compared against the proposed MotoSafety framework.}
\label{tab:model_performance}
\setlength{\tabcolsep}{2 pt}
\begin{tabular}{l l c c l l}
\toprule
\textbf{Model} & \textbf{Acc. (\%)} & \textbf{Prec.} & \textbf{Rec.} & \textbf{F1-Score (\%)} & \textbf{AUC (\%)} \\
& Mean $\pm$ Std & Result & Result & Mean $\pm$ Std & Mean $\pm$ Std \\
\midrule
RF & $84.93 \pm 0.16^{\ast\ast\ast}$ & 85.24 & 84.87 & $85.02 \pm 0.23^{\ast\ast\ast}$ & $93.68 \pm 0.31^{\ast\ast\ast}$ \\
CNN & $90.19 \pm 0.11^{\ast\ast\ast}$ & 93.82 & 80.68 & $86.74 \pm 0.21^{\ast\ast\ast}$ & $97.62 \pm 0.22^{\ast\ast\ast}$ \\
RNN & $90.94 \pm 0.12^{\ast\ast\ast}$ & 83.41 & 96.38 & $89.43 \pm 0.32^{\ast\ast\ast}$ & $97.27 \pm 0.34^{\ast\ast\ast}$ \\
LLM4TS & $90.06 \pm 0.14^{\ast\ast\ast}$ & 90.58 & 83.23 & $86.81 \pm 0.31^{\ast\ast\ast}$ & $97.31 \pm 0.42^{\ast\ast\ast}$ \\
Informer & $93.61 \pm 0.07^{\ast\ast\ast}$ & 89.72 & 94.88 & $92.21 \pm 0.22^{\ast\ast}$ & $99.12 \pm 0.21^{\ast\ast}$ \\
TST & $93.86 \pm 0.09^{\ast\ast\ast}$ & 93.67 & 90.74 & $92.18 \pm 0.29^{\ast\ast}$ & $99.23 \pm 0.28^{\ast}$ \\
TimesNet & $94.06 \pm 0.05^{\ast\ast\ast}$ & 89.23 & 96.69 & $92.77 \pm 0.38^{\ast\ast}$ & $99.18 \pm 0.24^{\ast}$ \\
iTransformer & $94.39 \pm 0.06^{\ast\ast}$ & 92.41 & 93.57 & $93.12 \pm 0.42^{\ast}$ & $99.22 \pm 0.32^{\ast}$ \\
PatchTST & $94.41 \pm 0.05^{\ast\ast}$ & 90.73 & 96.18 & $93.28 \pm 0.24^{\ast}$ & $99.31 \pm 0.22$ \\
Time-LLM & $94.46 \pm 0.07^{\ast}$ & 93.12 & 93.79 & $93.41 \pm 0.33^{\ast}$ & $99.24 \pm 0.31^{\ast}$ \\

\textbf{MotoSafety} & $\mathbf{94.97 \pm 0.08}$ & $\mathbf{93.80}$ & $\mathbf{93.90}$ & $\mathbf{93.70 \pm 0.20}$ & $\mathbf{99.33 \pm 0.30}$ \\
\bottomrule
\end{tabular}
\end{minipage}
\end{table}

\subsection{Forecasting Horizon Analysis}
\label{sec:lead_time_analysis}

To characterize how forecasting accuracy varies with prediction horizon, Table~\ref{tab:forecasting_by_horizon} reports MSE and MAE for the kinematic-state forecasting task (Section~\ref{sec:math_model}) across horizons up to 7.2s. Error increases gradually with horizon (MSE from 0.033 at $H=96$ to 0.045 at $H=720$), indicating the model retains meaningful predictive signal even at longer horizons rather than degrading sharply. We note this result characterizes future kinematic-state prediction and is reported separately from the same-window collision classification task described in Section~\ref{sec:Dataset_Details}; the two tasks use different targets and should not be read as a single fused collision-lead-time result.

\begin{table}
\centering
\scriptsize
\caption{Forecasting performance by prediction horizon (mean over 5 runs)}
\label{tab:forecasting_by_horizon}
\begin{tabular}{lccc}
\toprule
\textbf{Horizon $H$} & \textbf{Time (s)} & \textbf{MSE} & \textbf{MAE} \\
\midrule
96 & 0.96 & 0.033 $\pm$ 0.005 & 0.082 $\pm$ 0.007 \\
192 & 1.92 & 0.037 $\pm$ 0.006 & 0.086 $\pm$ 0.008 \\
336 & 3.36 & 0.041 $\pm$ 0.006 & 0.097 $\pm$ 0.010 \\
720 & 7.20 & 0.045 $\pm$ 0.007 & 0.111 $\pm$ 0.012 \\
\bottomrule
\end{tabular}
\end{table}

\subsection{Downstream Task: Impact of TP on Model Performance}
\label{sec:DownstreamTask}

\begin{table}[t]
\centering
\scriptsize
\caption{Performance comparison of MotoSafety showing improvement from adding TP features. Significance markers indicate improvement over the Without TP baseline: $^{***}p < 0.001$}
\label{tab:predicted_tp_results}
\begin{tabular}{lll}
\midrule
\textbf{Model} & \textbf{Accuracy (\%)} & \textbf{AUC (\%)} \\
\midrule
Without TP Feature & $94.09 \pm 0.04$ & $99.10 \pm 0.01$ \\
With Predicted TP Feature & $94.82 \pm 0.02^{***}$ & $99.24 \pm 0.01^{***}$ \\
With Ground Truth TP Feature & $\mathbf{94.97 \pm 0.01^{***}}$ & $\mathbf{99.33 \pm 0.01^{***}}$ \\
\bottomrule
\end{tabular}
\end{table}

In practical deployment, direct and accurate measurement of ground truth TP labels, denoted \( TP_{gt} \), may be infeasible due to sensor limitations or real-time constraints. In the proposed data set, the TP characteristic is a categorical variable representing three distinct TP conditions: HTP = 0, LTP = 1, NTP = 2. To address missing or unavailable GT TP labels, we employed a Time Series Transformer (TST)~\citep{zerveas2020transformerbasedframeworkmultivariatetime} model to predict these categorical TP labels from the multivariate time series input data. The TST achieved 89.26\% accuracy for 3-class TP prediction (0=HTP, 1=LTP, 2=NTP). Table~\ref{tab:tst_confusion} shows the confusion matrix (raw counts) per-class performance. 

\begin{table}
\centering
\scriptsize
\caption{Confusion matrix (raw counts) for TST-based TP prediction.}
\label{tab:tst_confusion}
\begin{tabular}{lccc}
\toprule
True / Predicted & 0: HTP & 1: LTP & 2: NTP \\
\midrule
0: HTP & 8878 & 229 & 163 \\
1: LTP & 76 & 7326 & 568 \\
2: NTP & 82 & 1661 & 6859 \\
\bottomrule
\end{tabular}
\end{table}

Let $\mathbf{X} = (\mathbf{x}_1, \mathbf{x}_2, \dots, \mathbf{x}_T) \in \mathbb{R}^{T \times D}$ represent the multivariate time series input from the PTW simulator, which includes all features except the explicit TP indicator. In this notation, $T$ refers to the sequence length, while $D = 63$ specifies the number of telemetry features. Each $\mathbf{x}_t \in \mathbb{R}^D$ captures the sensor states at the $t$-th time instant. The TST model is then expressed as:

\begin{equation}
TP_{pred} = f_{TST}(\mathbf{X}),
\label{eq:tst_pred}
\end{equation}
where \( TP_{pred} \in \{0,1,2\} \) is the predicted TP label. The predicted TP feature \( TP_{pred} \), along with all other relevant sensor and behavioral features of the data set, forms the complete input to our collision risk prediction model, MotoSafety. Formally, the model can be expressed as:
\begin{equation}
\hat{y} = g_{MotoSafety}(\mathbf{X}, TP),
\label{eq:motors_model}
\end{equation}
where \( TP \) is the ground truth \( TP_{gt} \) or predicted \( TP_{pred} \), and \(\hat{y}\) is the predicted safety risk label for the rider. We evaluated the robustness of model by comparing the predictions when using:
\begin{align}
\hat{y}_{gt} &= g_{MotoSafety}(\mathbf{X}, TP_{gt}), \label{eq:model_gt}\\
\hat{y}_{pred} &= g_{MotoSafety}(\mathbf{X}, TP_{pred}). \label{eq:model_pred}
\end{align}

The results presented in Table~\ref{tab:predicted_tp_results} highlight the critical role of TP in predicting collision risk. While the baseline model \( f_{\text{Motosafety}}(X)\) achieves an accuracy 94.09\% using only PTW simulation data without TP feature. The integration of predicted TP features \( f_{\text{Motosafety}}(X \oplus TP_{pred})\) improved this to 94.82\%, nearly matching the 94.97\% Oracle performance \( f_{\text{Motosafety}}(X \oplus TP_{gt})\). This performance gain shows that $TP_{pred}$ captures latent psychological context, such as rider stress, which is not fully represented by telemetry alone. The marginal 0.15\% gap between the predicted (94.82\%) and Oracle (94.97\%) results validates MotoSafety for real-world deployment where ground truth states are not available.

\subsection{Model Performance and Reliability}

\subsubsection{Confusion Matrix Analysis}

\begin{figure}
\centering
\includegraphics[width=0.42\textwidth]{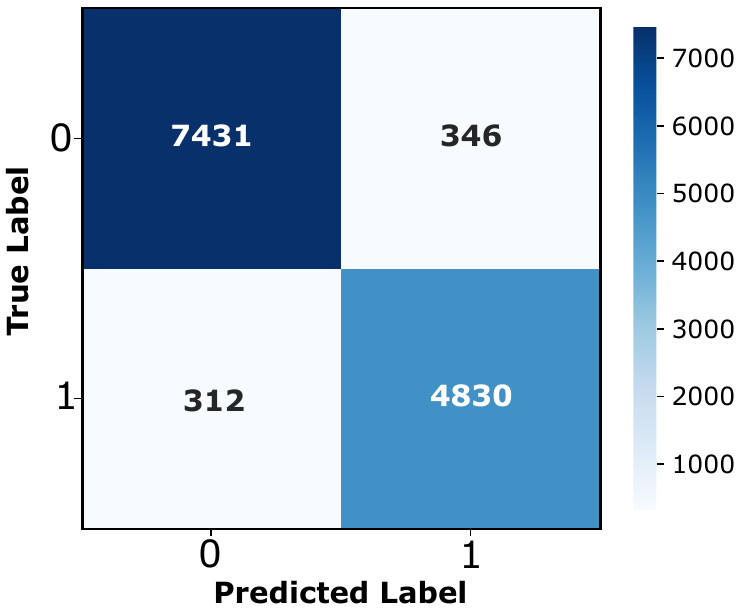}
\caption{Confusion matrix of MotoSafety predictions on the main PTW simulator dataset under TP.}
\label{fig:confusion_matrix}
\end{figure}

The confusion matrix provides a detailed summary of MotoSafety model classification outcomes for collision and non-collision cases. Figure~\ref{fig:confusion_matrix} shows the distribution of accurate and inaccurate predictions.

\subsubsection{Calibration Curve and Risk Scores}

\begin{figure}
    \centering
    \begin{minipage}{0.42\linewidth}
        \centering
        \includegraphics[width=\linewidth]{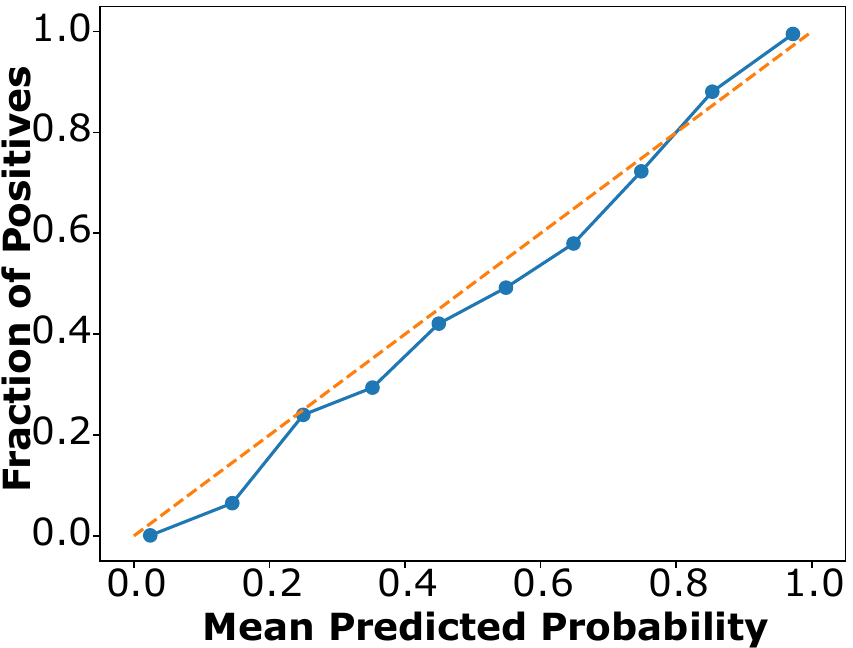}
        \caption{Calibration curve.}
        \label{fig:calibration_curve}
    \end{minipage}
    \hfill
    \begin{minipage}{0.45\linewidth}
        \centering
        \includegraphics[width=\linewidth]{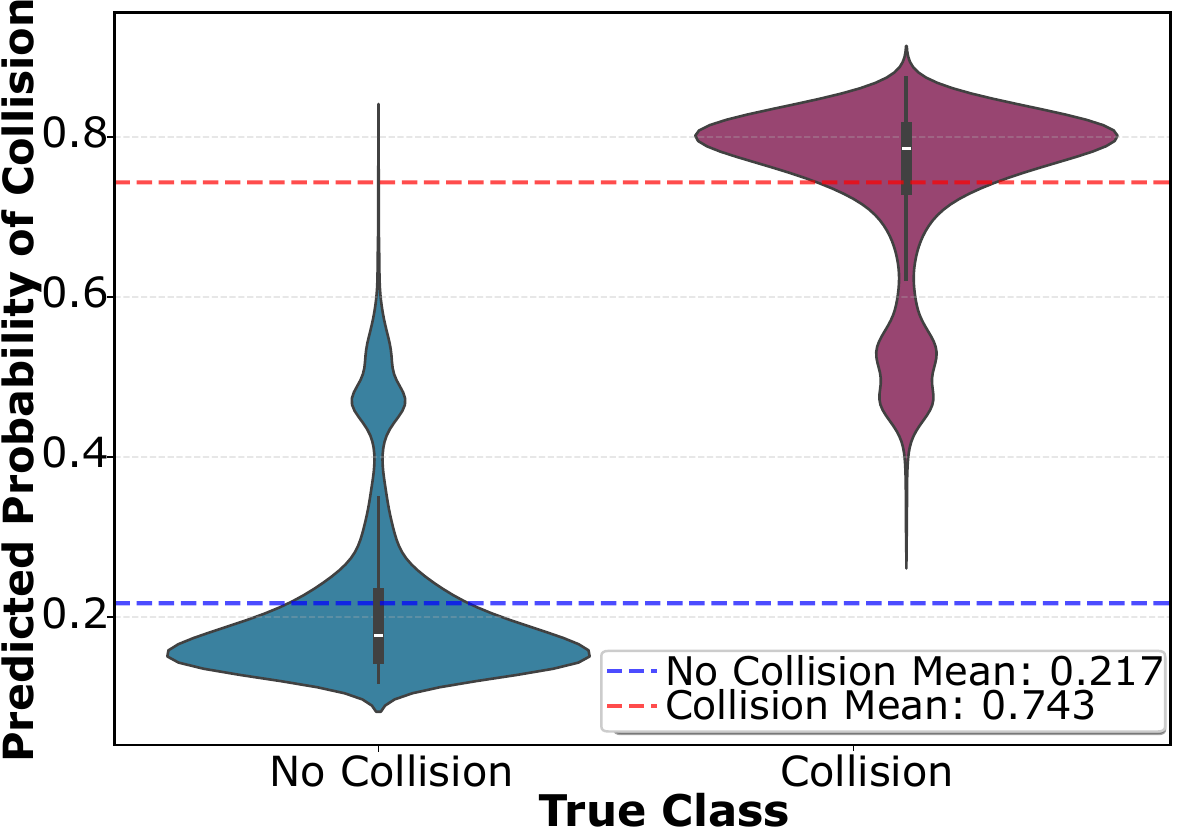}
        \caption{Distribution of predicted collision risk.}
        \label{fig:violin_risk}
    \end{minipage}
\end{figure}

The MST calibration curve on the primary PTW simulator dataset under TP (Fig.~\ref{fig:calibration_curve}) shows that predicted collision probabilities closely match observed event frequencies, indicating statistically reliable and well-calibrated outputs. In safety-critical contexts, such calibration enables risk-aware adaptive interventions rather than reliance on fixed thresholds. The distribution of predicted risk scores by true outcome (Fig.~\ref{fig:violin_risk}) further illustrates clear separation: collision events consistently receive higher predicted probabilities, while non-collision samples remain concentrated at lower values. This separation reinforces both discriminative ability and calibration quality, supporting the potential for real-world deployment.

\subsection{Architecture Transferability Across Safety Domains}
\label{sec:Generalizability}

\begin{table*}[t]
\centering
\scriptsize
\caption{Cross-Safety-Domain Evaluation of MotoSafety Accuracy (\%) and Prior Work Accuracy (\%). MST: MotoSafety. }
\label{tab:cross_dataset}
\renewcommand{\arraystretch}{1.0}
\setlength{\tabcolsep}{2pt} 
\begin{tabular}{@{} p{3.5cm} p{2.7 cm} p{2 cm} p{2.3cm} p{1.1cm} p{1cm} p{3.5cm} @{}}
\toprule
\textbf{Dataset} & \textbf{Details} & \textbf{Safety Domain} & \textbf{Application} & \textbf{Country} & \textbf{MST} & \textbf{Prior Work Accuracy} \\
\midrule
Proposed Dataset & PTW Simulator with GT TP  & PTW Safety & Collision Prediction & India & \textbf{94.97} & -- \\

\cite{DARUS33012023} & PTW High-Fidelity Simulator & PTW Safety & Collision Prediction & Germany & \textbf{99.49} & 91.0 [RF, GB~\citep{Rodegast2024}] \\

\cite{BOUBEZOUL2019103828} & PTW Real Time Fall Event Data & PTW Safety & Fall Detection & France & \textbf{96.10} & 91.59 [DT~\citep{fi15100333}] \\

\cite{har240} & Human Activity Recognition & Human Safety &  Activity Recognition & Italy & \textbf{97.66} & 83.35 [K-means + NB~\citep{Ismi2016KmeansCB}] \\

\cite{physical_therapy_exercises_730} & Wearable-Sensor Movement & Human Clinical Safety & Exercise Quality Assessment & Turkey & \textbf{99.65} & 88.65[MTMM-DTW~\citep{YURTMAN2014189}] \\

\bottomrule
\end{tabular}
\end{table*}

Note that each dataset below is used to train and evaluate a separate MotoSafety instance from scratch (i.e., no weights are transferred from the PTW model); this experiment therefore evaluates the transferability of the architecture to other safety-critical sequence classification tasks, rather than knowledge transfer from the PTW domain itself.

To evaluate the versatility of the MotoSafety framework beyond motorcycle simulators, we performed cross-dataset evaluations across distinct safety-critical domains as shown in Table~\ref{tab:cross_dataset}. In the PTW Safety domain, MotoSafety achieved 99.49\% accuracy on the high-fidelity German simulator dataset \citep{DARUS33012023} for collision prediction, outperforming traditional ensemble methods (RF/GB) by 8.49\%. Similarly, on real-world fall event data \citep{BOUBEZOUL2019103828}, achieved improved 96.10\% accuracy, a significant improvement over the 91.59\% reported for Decision Tree (DT) benchmarks \citep{fi15100333}.
Beyond vehicle dynamics, the model exhibited exceptional robustness in Human and Clinical Safety. On the UCI HAR human activity recognition dataset \citep{har240}, MotoSafety surpassed Naive Bayes (83.35\%) by 14.31\%, achieving 97.66\% accuracy. Most notably, in clinical exercise quality assessment \citep{physical_therapy_exercises_730}, the framework achieved near-perfect classification (99.65\%), marking an 11\% gain over DTW-based clinical algorithms (88.65\%) \citep{YURTMAN2014189}. These results demonstrate that the MotoSafety architecture is not overfit to PTW-simulator-specific artifacts and performs competitively when retrained on diverse safety-critical sequence classification tasks, supporting the architecture's applicability beyond its original design domain.

\subsection{Ablation Study}

\label{sec:Ablation}

We evaluated individual contributions of MotoSafety components through a comprehensive ablation study shown in Table~\ref{tab:ablation}. The study was carried out on PTW simulation data under TP. The complete model attains an accuracy of 94.97\%. The removal of BiLSTM and MHA resulted in significant performance degradation of 4.55\% and 4.19\%, respectively, highlighting their role in capturing long-range dependencies. Similarly, the exclusion of the CNN branch and Gated Fusion mechanism led to accuracy drops of 2.31\% and 2.07\%, underscoring the necessity of multi-scale feature extraction. Furthermore, removing the SE block decreases accuracy by 3.73\%, validating its effectiveness in channel-wise feature recalibration. Collectively, these findings justify the necessity of each integrated module to maintain high predictive precision.

\begin{table}
\centering
\scriptsize
\caption{Ablation Study of MotoSafety Components.}
\label{tab:ablation}
\setlength{\tabcolsep}{1pt}
\renewcommand{\arraystretch}{1.1}
\begin{tabular}{@{}p{3cm}rrl@{}}
\midrule
\textbf{Variant} & \textbf{Accuracy (\%)} & \textbf{$\Delta$ Acc.} & \\
\midrule
Full Model & \textbf{94.97} & -- & \\
No BiLSTM & 90.42 & $-4.55$ & \\
No SE Block & 91.24 & $-3.73$ & \\
No Attention & 90.78 & $-4.19$ & \\
No CNN Branch & 92.66 & $-2.31$ & \\
No Gated Fusion & 92.90 & $-2.07$ & \\
\bottomrule
\end{tabular}
\end{table}

\begin{table}
\centering
\scriptsize
\caption{Comparison of MotoSafety with TIP against standard fixed temporal-pooling strategies (mean $\pm$ std over 5 runs).}
\label{tab:tip_ablation}
\begin{tabular}{lcc}
\toprule
\textbf{Pooling Strategy} & \textbf{Accuracy (\%)} & \textbf{AUC (\%)} \\
\midrule
Mean-pooling & 92.23 $\pm$ 0.15 & 98.80 $\pm$ 0.22 \\
Max-pooling & 93.41 $\pm$ 0.12 & 99.11 $\pm$ 0.18 \\
Last-timestep & 92.87 $\pm$ 0.14 & 98.97 $\pm$ 0.20 \\
\midrule
\textbf{TIP} & \textbf{94.97 $\pm$ 0.08} & \textbf{99.33 $\pm$ 0.30} \\
\bottomrule
\end{tabular}
\end{table}

To further isolate the effect of TIP, we compare it against three standard temporal-pooling strategies—mean-pooling, max-pooling, and last-timestep extraction—under an otherwise identical architecture. As shown in Table~\ref{tab:tip_ablation}, MotoSafety with TIP achieves 94.97\% accuracy, outperforming mean-pooling by 2.74\%, max-pooling by 1.56\%, and last-timestep by 2.10\%. This indicates that the performance gain stems from the learned, content-aware weighting rather than from the surrounding architecture alone.

\subsection{Real-World Feature Feasibility}
\label{sec:real_world_features}

To assess real-world deployment feasibility, we identified that 21 of the 64 features used in this study are available on standard motorcycles via low-cost IMU+GPS (e.g., speed, acceleration, yaw rate, position, brake force, throttle, steering angle). The remaining features (e.g., precise gap metrics, rule violations) are simulator-only. As shown in Table~\ref{tab:bike_features}, accuracy increases with more features, achieving 93.91\% with 21 features (compared to 94.97\% with all 64 features), indicating that real-world deployment is feasible with onboard sensors. Testing on actual hardware in real traffic conditions remains essential and is therefore deferred to future work, as discussed in Section~\ref{sec:LimitationsFuture}.

\begin{table}
\centering
\scriptsize
\caption{MotoSafety accuracy with real-bike features (IMU+GPS).}
\label{tab:bike_features}
\begin{tabular}{lcc}
\toprule
\textbf{Features} & \textbf{Accuracy (\%)} \\
\midrule
3 & 80.22 \\
5 & 83.11 \\
7 & 88.78 \\
9 & 90.16 \\
21 & 93.91 \\
All 64 (simulator) & 94.97 \\
\bottomrule
\end{tabular}
\end{table}

\subsection{Model Complexity and Edge-AI Deployability}
\label{sec:model_complexity}

The MotoSafety model is designed for real-time deployment in resource-constrained environments. Latency was measured on an Intel Core i7 CPU (2.8 GHz) with batch size 32, PyTorch 2.0, CPU-only inference, averaged over 1000 batches. 
With only 1.15M parameters and a 4.38 MB memory footprint (Table~\ref{tab:combined_metrics}A), it maintains a low computational profile without sacrificing accuracy. Benchmarking on our PTW simulation dataset reveals that MotoSafety achieves a per-sample inference latency of 0.135 ms, outperforming all baseline and SOTA architectures (Table~\ref{tab:combined_metrics}B). Specifically, MotoSafety is \textbf{5.3$\times$, 9.5$\times$, 21.9$\times$, and 71.9$\times$ faster} than PatchTST, Time-LLM, TimesNet, and LLM4TS (GPT-2), respectively. Even when accounting for TST preprocessing (36.88ms overhead) to predict TP labels when ground truth is unavailable, MotoSafety (37.01ms) remains faster than iTransformer (37.47ms), PatchTST (37.60ms), Time-LLM (38.16ms), TimesNet (39.84ms), and LLM4TS (46.59ms). This ultra-low latency ensures near-instantaneous risk assessment, allowing maximum time for emergency interventions. These efficiency metrics satisfy the stringent requirements for real-time deployment on low-cost on-board units (OBUs). Unlike GPU-dependent Transformer baselines, complexity of MotoSafety is $\mathcal{O}(L)$. This balance of efficiency and performance makes it a viable candidate for large-scale ITS deployment in developing regions, providing an accessible AI-driven safety net for vulnerable riders.

\begin{table}
\centering
\scriptsize
\caption{MotoSafety Efficiency Metrics and Comparative Latency.}
\label{tab:combined_metrics}
\setlength{\tabcolsep}{1pt} 
\renewcommand{\arraystretch}{1}
\begin{tabular}{@{}p{2.6cm}p{2cm}p{1.8cm}p{1.8cm}@{}}
\toprule
\multicolumn{4}{@{}l}{\textbf{A. MotoSafety Efficiency \& Deployability}} \\
\midrule
\textbf{Metric} & \textbf{Value} & \multicolumn{2}{l}{\textbf{Significance}} \\
\midrule
Total Parameters & 1,149,048 & \multicolumn{2}{l}{Low Complexity} \\
Model Size & 4.38 MB & \multicolumn{2}{l}{Edge-ready Storage} \\
Throughput & $> 7,400$ s/sec & \multicolumn{2}{l}{High-speed Inference} \\
Arch. Complexity & $\mathcal{O}(L)$ & \multicolumn{2}{l}{Linear Scalability} \\
Edge-AI Deployability & Yes & \multicolumn{2}{l}{Suitable for wearable} \\
Deployment Regions & Global & \multicolumn{2}{l}{Operate in low-resource} \\
\midrule
\multicolumn{4}{@{}l}{\textbf{B. Latency Comparison (PTW Simulator Dataset Under TP)}} \\
\midrule
\textbf{Model (Year)} & \textbf{Latency (ms)} & \textbf{Speed Gap} & \textbf{Parameters} \\
\midrule
MotoSafety & 0.135 & 1.0× & 1,149,048 \\
iTransformer (2024) & 0.587 & 4.3× $\downarrow$ & 275,714 \\
PatchTST (2023) & 0.720 & 5.3× $\downarrow$ & 187,330 \\
Time-LLM (2024) & 1.280 & 9.5× $\downarrow$ & 3,183,618 \\
TimesNet (2023) & 2.960 & 21.9× $\downarrow$ & 4,708,226 \\
LLM4TS (2025) & 9.710 & 71.9× $\downarrow$ & 124,506,626 \\
\bottomrule
\end{tabular}
\end{table}

\section{Conclusion}
\label{sec:conclusion}

In this work, we investigated how TP affects PTW riders and contributes to collision risk. To fill this gap, we collected a large-scale dataset comprising over 129,209 feature windows extracted via sliding window segmentation from 153 riding sessions (51 riders × 3 TP conditions) under no, low, and high TP conditions. Each sequence includes 64 features covering vehicle dynamics, control inputs, proximity measures, temporal context, and behavioral violations.

Building on this dataset, we proposed MotoSafety, a novel deep learning architecture grounded in the LTI principle. MotoSafety achieves 94.97\% accuracy and 99.33\% ROC AUC for collision risk assessment, outperforming ten baselines including TimesNet, PatchTST, iTransformer, Time-LLM, and LLM4TS. For long-term forecasting, it achieves 0.039 MSE and 0.094 MAE (4.4× lower error than Time-LLM and iTransformer). With only 1.15 million parameters and 0.135 ms inference latency, the model is 21.9$\times$ and 71.9$\times$ faster than TimesNet and LLM4TS.

Our findings show that explicit TP prediction provides a critical inductive bias, improving collision accuracy from 94.09\% to 94.97\% (ground truth) and 94.82\% with predicted TP. Furthermore, using only 21 real-world features (available via low-cost IMU+GPS), MotoSafety achieves 93.91\% accuracy, indicating practical deployment potential; real-world hardware validation remains future work. Beyond PTW safety, the MotoSafety architecture shows improved transferability when retrained on human activity recognition (97.66\%) and clinical exercise monitoring (99.65\%) datasets.

The model's small size (1.15M parameters, 4.38 MB) and low latency (0.135 ms) make it well-suited for future edge deployment on handlebar-mounted devices or smart helmets as a potential edge-deployable black box for two-wheelers, enabling real-time collision risk alerts without relying on cloud connectivity.

\subsection{Limitations and Future Work}
\label{sec:LimitationsFuture}

We acknowledge several limitations in the present study. First, our participant pool consisted solely of male riders, which aligns with Indian fatality statistics showing that males account for 85.2–87.3\% of PTW deaths. However, this limits the generalizability of our findings to female riders, since perceptions of TP and riding behavior may vary between genders. Future studies should include female riders to systematically examine whether the observed relationships hold across genders. Second, the data were gathered in a controlled simulator setting, which may not completely capture real-world factors like rider fatigue, changing weather, or social pressures. Real-world TP collision data cannot be collected safely, so simulators provide the only ethical controlled setting. To validate in real conditions, we plan phased testing: closed-course trials followed by naturalistic data collection through commercial riding platforms. Third, we did not record physiological stress markers, such as heart rate variability or electrodermal activity. Behavioral markers such as speed variability and control inputs were used as indicators of cognitive load; however, the absence of physiological validation limits direct assessment of the underlying stress response. Future work should incorporate physiological measures to complement behavioral markers. Fourth, the proposed safety system requires prospective validation in real-world operational settings before practical deployment. Hardware validation on handlebar-mounted devices or smart helmets remains future work.

Future research will assess MotoSafety across varied rider populations and road conditions using both simulator-based and naturalistic datasets. We plan to incorporate multimodal signals—including physiological, behavioral, and contextual data—to improve robustness. Adaptive ITS interventions, such as haptic feedback, throttle modulation, and context-sensitive alerts, will be explored to mitigate crash risk and manage kinetic energy transfer. Transfer learning and domain adaptation techniques will also be applied to ensure the framework is scalable across different vehicle types, regions, and cultural contexts.

\section*{Data and Code Availability}
    
The data and code supporting this study will be made available by the corresponding author upon reasonable request after publication.



\section*{Funding}

This research received funding from the IIT Indore Young Faculty Research Catalyzing Grant (YFRCG) Scheme under Project ID: IITI/YFRCG/2023-24/01.

\section*{Acknowledgments}

The authors gratefully acknowledge all the volunteers who took part in this study. We also extend our sincere thanks to Manvendra T. for his valuable assistance with data collection.
    
\section*{Declaration of Competing Interest}

The authors declare that they have no known competing financial interests or personal relationships that could have influenced the work reported in this study.




  \bibliographystyle{elsarticle-harv} 

 \bibliography{cas-refs}

    
    
    \end{document}